\documentclass{article}
\usepackage{acm_arxiv}
\usepackage{booktabs} % For formal tables

\usepackage[ruled]{algorithm2e} % For algorithms

\SetAlFnt{\small}
\SetAlCapFnt{\small}
\SetAlCapNameFnt{\small}
\SetAlCapHSkip{0pt}

\usepackage{caption}
\usepackage{subcaption}
\usepackage{graphicx}
\usepackage{amsmath}
\usepackage{amsthm}
\usepackage{booktabs}
\newcommand{\mc}[1]{\mathcal{#1}}

\newcommand{\vts}[1]{\lvert #1 \rvert}

\newcommand\Tstrut{\rule{0pt}{2.6ex}}         % = `top' strut
\newcommand\Bstrut{\rule[-1.3ex]{0pt}{0pt}}   % = `bottom' strut
\makeatletter
\newcommand\footnoteref[1]{\protected@xdef\@thefnmark{\ref{#1}}\@footnotemark}
\makeatother

\newcommand{\shorteq}{%
  \settowidth{\@tempdima}{-}% Width of hyphen
  \resizebox{\@tempdima}{\height}{=}%
}

\usepackage{amssymb,fge}

\usepackage{amsmath, amssymb}
\usepackage{subcaption}
\usepackage[utf8]{inputenc}
\usepackage[T1]{fontenc}
\usepackage[font=small,belowskip=-1.5pt]{caption} % This causes problems while compiling to latex    
\usepackage{array}
\usepackage{graphicx}
\usepackage{amsfonts}
\usepackage{soul}
\usepackage{hhline}
\usepackage{multirow, makecell}
\usepackage{float}
\usepackage{booktabs}

\usepackage{amsthm}
\usepackage{color}
\usepackage{transparent}
\usepackage{url}
\usepackage{footmisc}
\usepackage{setspace}
\usepackage{textcomp}
\usepackage{mathtools}

\DeclareMathOperator*{\argmax}{arg\,max}

\theoremstyle{plain}

\mathchardef\mhyphen="2D

\begin{document}

% Title portion
\title{BoMuDANet: Unsupervised Adaptation for Visual Scene Understanding in Unstructured Driving Environments}

 \author{
    Divya Kothandaraman,
    Rohan Chandra,
    Dinesh Manocha \\
    University of Maryland, College Park \\
    {\small Tech Report, Code, and Video at \url{https://gamma.umd.edu/bomuda}}}

\maketitle
\begin{abstract}

We present an unsupervised adaptation approach for visual scene understanding in unstructured traffic environments. Our method is designed for unstructured real-world scenarios with dense and heterogeneous traffic consisting of cars, trucks, two-and three-wheelers, and pedestrians.  We describe a new semantic segmentation technique based on unsupervised domain adaptation (DA), that can identify the class or category of each region in RGB images or videos. We also present a novel self-training algorithm (Alt-Inc) for multi-source DA that improves the accuracy. Our overall approach is a deep learning-based technique and consists of an unsupervised neural network that achieves 87.18\% accuracy on the challenging India Driving Dataset. Our method works well on roads that may not be well-marked or may include dirt, unidentifiable debris, potholes, etc.  A key aspect of our approach is that it can also identify objects that are encountered by the model for the fist time during the testing phase.  We compare our method against the state-of-the-art methods and show an improvement of $5.17\%$ \textminus \ $42.9\%$. Furthermore, we also conduct user studies that qualitatively validate the improvements in visual scene understanding of unstructured driving environments.  
\end{abstract}

\section{Introduction}
\label{sec: introduction}

Visual scene understanding is a key component of perception systems in autonomous vehicles (AVs) that deals with tracking, prediction, object detection, classification, localization, and semantic segmentation \cite{yu2017dilated,schwarting2018planning}. These systems are responsible for understanding or interpreting the environment for safe and efficient navigation and collision avoidance. There has been considerable work on developing systems that are now deployed in 
the current generation of AV technologies~\cite{tesla2020}. However, most of the scene understanding algorithms and systems have been developed for highly controlled or structured environments.  This includes sparse or homogeneous  traffic, well-structured roads with clear lane marking, absence of debris, potholes or unidentifiable objects etc. Many times, the AVs need to operate in unstructured scenarios that consist of heterogeneous traffic with cars, trucks, bicycles or pedestrians and there is less adherence to traffic rules or regulations. Furthermore, roads are not well-marked or may include dirt or unidentifiable debris. There are still many challenges in terms of developing robust perception systems for such unstructured environments.

%with dirt real word environments structured, especially in developing nations such as India where lane markings are often missing and traffic is not regularized. Current systems for visual scene understanding have not been tested in such unstructured environments. In this work, we propose a new perception algorithm for visual scene understanding in dense and unstructured traffic environments.
\begin{figure}[t]
    \centering
    \includegraphics[width=\columnwidth, height=5.68cm]{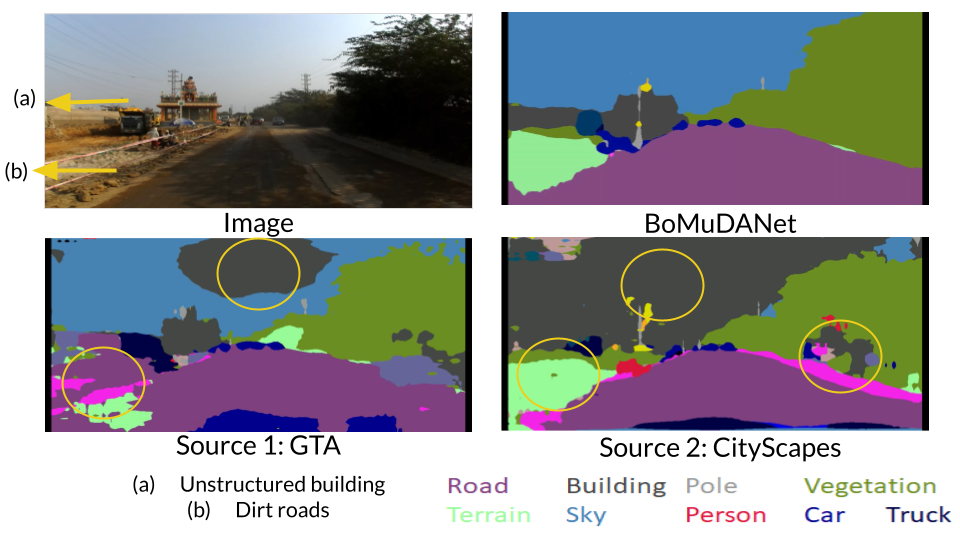}
    \caption{We present a novel unsupervised deep learning-based approach called BoMuDANet for visual scene understanding in unconstrained and unstructured traffic environments. In this example, we demonstrate the benefits of BoMuDANet on images taken from the challenging IDD dataset. BoMuDANet accurately segments out dirt roads as terrains as well as a building, while preserving its shape. In contrast, the single source baselines (GTA/CityScapes) do not identify dirt and unstructured roads well, misclassify parts of sky as building, and fail to capture the shape of the unstructured building. BoMuDANet benefits from its ability to selectively distil information from various sources by iterative self-training, in addition to exploiting a chosen best source via domain adaptation.}
    \label{fig: cover}
    %\vspace{-20pt}
\end{figure}

Recent developments in deep learning~\cite{chen2017deeplab,yu2017dilated} have resulted in significant advances in visual scene understanding~\cite{hofmarcher2019visual,xiao2012basic} in structured traffic environments~\cite{yu2020bdd100k,cs}. However, they do not work well in unstructured or unconstrained environments like  India Driving Dataset~\cite{varma2019idd} (see Figure~\ref{fig: cover}). This is mainly  due to lack of good  datasets that consists of unstructured traffic scenes or outlier objects. For example, current traffic datasets used for training may not have vehicles such as  auto-rickshaws~\cite{campbell2010autonomous}. An approach that can overcome these challenges is domain adaptation (DA)~\cite{DA}, a transfer learning technique that takes advantage of the availability of large scale annotated data in a different domain called the `source' domain to perform a task on the `target' domain, for which data is typically scarce. More specifically, DA can leverage many available large-scale structured traffic datasets such as CityScapes~\cite{cs}, Berkeley Deep Drive~\cite{yu2020bdd100k}, GTA~\cite{gta}, and SynScapes\cite{wrenninge2018synscapes} (source domains) to learn robust feature representations in unstructured environments.

\paragraph{Main Contributions:} We present a new deep neural network called BoMuDANet for visual scene understanding in unstructured traffic environments. Our approach consists of a semantic segmentation technique based on unsupervised domain adaptation. BoMuDANet includes two novel components:

\begin{enumerate}

\item Unconstrained traffic environments are highly heterogeneous (wide range of object classes)~\cite{chandra2019traphic}; consequently, using only one source dataset (single source DA) is not sufficient in terms of providing the network with adequate information for optimal performance on the complex target domain. BoMuDANet benefits from its ability to selectively extract relevant knowledge across different and widely available structured environment datasets~\cite{yu2020bdd100k,cs}. Moreover, we 
%offers an advantage where one can theoretically
perform multi-source DA~\cite{zhao2019multisemantic} by alternating between adaptation from a selected source and knowledge distillation from the remaining sources. Based on the classical EM algorithm in statistical pattern recognition, BoMuDANet performs repeated rounds of training, alternating between adaptation and distillation to improve performance in each step. We present the Alt-Inc algorithm in Section~\ref{subsec: multi-source_DA}.

\item Unconstrained traffic scenes may contain objects that are typically non-existent in current structured environment source domain datasets. Our approach in BoMuDANet uses a simple pseudo-labeling strategy (Section~\ref{sec: boundless}) for handling unknown objects encountered for the first time during the testing phase. 
The final probability predictions of the Alt-Inc algorithm are directly used to assign proxy labels to unknown object classes depending on their  similarity to known objects classes in the training dataset. Our approach helps BoMuDANet detect new objects that are common in unstructured driving environments.
\end{enumerate}

We have evaluated our approach extensively using the Indian Driving Dataset (IDD), CityScapes, Berkeley DeepDrive, GTA V, and the Synscapes datasets. In unstructured environments (IDD as the target dataset), we show that our unsupervised approach outperforms other unsupervised SOTA benchmarks by $5.17\%-42.9\%$. In structured environments (CityScapes as the target dataset), we show that our method outperforms other multi-source DA methods by $12.70\%-90.13\%$. We have performed extensive ablation experiments to highlight the benefits of our approach. Moreover, we also perform a user study to highlight the qualitative benefits of our approach. Overall, ours is the first unsupervised domain adaptation method for handling unstructured traffic environments. %Finally, we perform ablation analyses to highlight the benefits of the different components in our algorithm.

\section{Prior Work}
\label{sec: related}

There is considerable work in domain adaptation (DA) for semantic segmentation and other perception tasks. While a detailed review of these methods is not within the scope of this paper, we briefly mention related work. 

\subsection{Semantic Segmentation}

Semantic segmentation is a pixel-level task, which involves assigning a label to each pixel in an image. The advent of deep learning has resulted in a many segmentation techniques for autonomous driving~\cite{hao2020brief,yu2017dilated,chen2017deeplab,takikawa2019gated,fu2019dual,guo2021survey}. However, these methods suffer from three issues: (i) the networks need to be trained in a supervised manner, thus there is a demand for large volumes of annotated data; (ii) current labeled datasets are limited to structured environments;  (ii) current learning methods do not scale well to unstructured environments.

\subsection{Unsupervised Domain Adaptation}

Domain adaptive semantic segmentation has been explored under three different machine learning paradigms that differ based on the underlying learning approach. At one end of the spectrum, fully supervised~\cite{effu} methods achieve higher accuracy on average, but are limited by the availability of annotated data. On the other end of the spectrum, unsupervised methods`\cite{pan2020unsupervised,vu2019advent,tsai2018learning,liu2020unsupervised,pan2020unsupervised,zheng2019unsupervised,zhang2020jointunsupervised,jaritz2020xmudaunsupervised,wang2020differentialunsupervised,yang2020adversarialunsupervised} 
benefit from the lack of dependence on any training data, but are outperformed by fully supervised or semi-supervised methods. Semi-supervised approaches~\cite{kim2020attract,li2020online,saito2019semi,qin2020opposite}
form a middle ground between the two paradigms. While there has been some work on using pseudo labels \cite{bucher2020buda,kothandaramandomain} for training DA models, they do not scale well in the presence of multiple sources.  

%Many recent approaches in unsupervised domain adaptation (UDA) involve adversarial training. \cite{tsai2018learning} uses multi-level domain adaptation, while \cite{vu2019advent} builds upon the former and maps probability maps to entropy. \cite{hoffman2016fcns} aligns domains at the pixel level, while \cite{hoffman2018cycada} uses cycleGAN to transform images from one domain to another followed by domain alignment. Other domain adaptation methods for urban scenes include \cite{chen2018road,zhang2017curriculum,wu2018dcan}. Past work \cite{choi2019pseudo,chang2019domain,bucher2020buda,das2018graph,zhang2019category} has also relied on pseudo-labeling for self-training UDA models. \divya{how do we improve over the past methods that use pseudo labels?}

% In the recent past, there has been a rapid growth in unsupervised domain adaptive semantic segmentation 
\begin{figure}[t]
    \centering
    \includegraphics[width = .7\columnwidth]{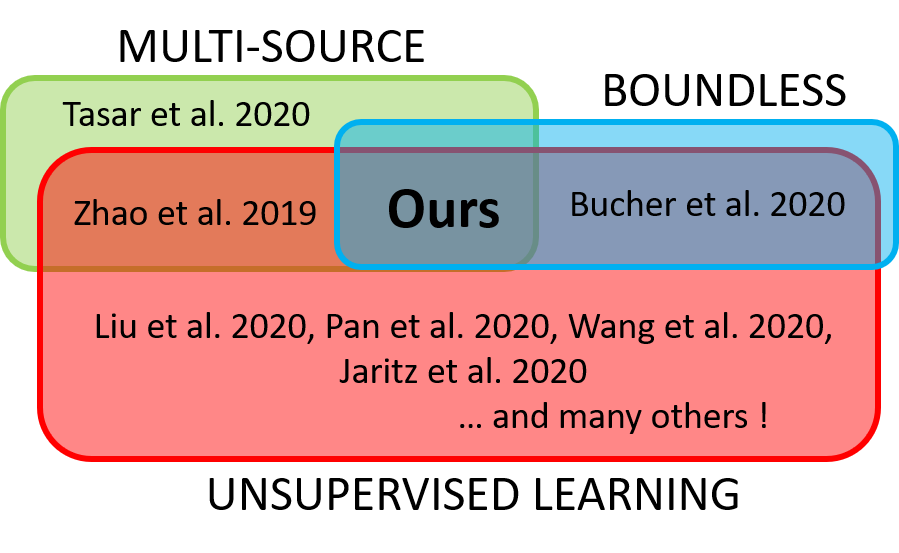}
    \caption{Extension to SOTA in domain adaptive semantic segmentation. Our approach is the first method to \textit{simultaneously} perform unsupervised multi-source boundless DA segmentation and can handle unstructured traffic environments.}
    
    \label{fig: venn_diagram}
    \vspace{-15pt}
\end{figure}
\begin{figure*}[t]
    \centering
    \includegraphics[width = \textwidth, height=6.8cm]{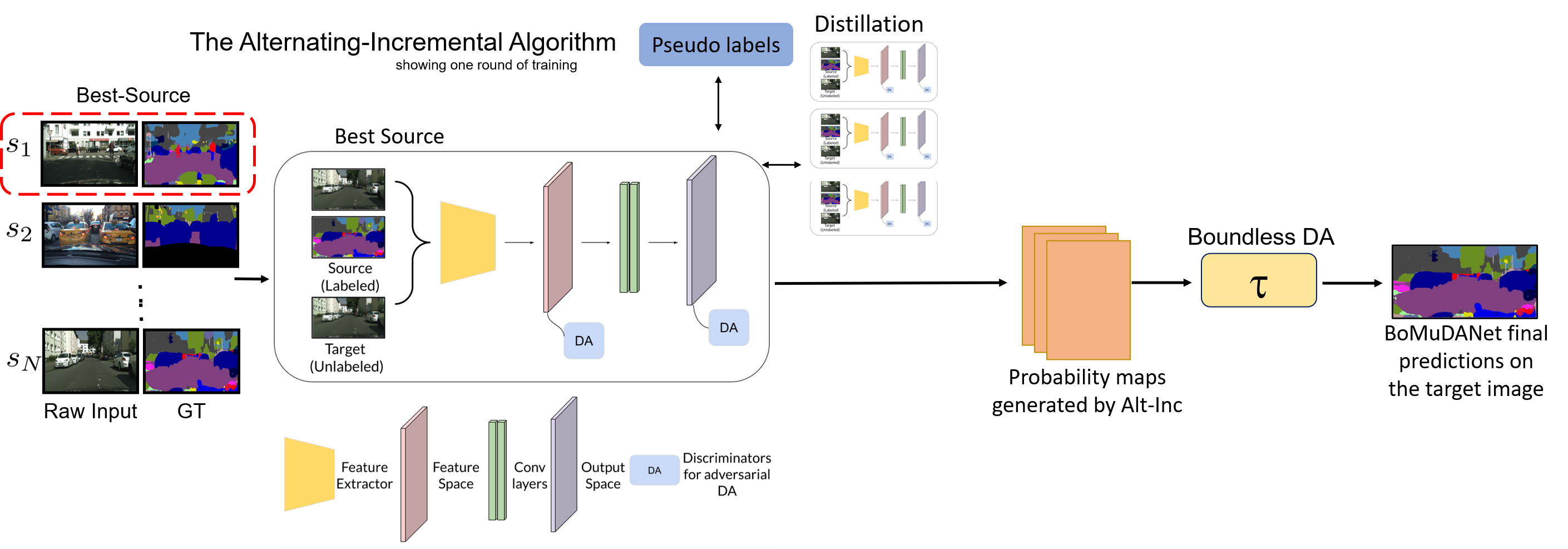}
    \caption{\textbf{Overview of BoMuDANet:} The input consists of $N$ sources ($s_1, s_2, \ldots, s_N$), from which the best-source is selected by the Alternating-Incremental algorithm (Section~\ref{subsec: multi-source_DA}). The Alt-Inc algorithm proceeds in an unsupervised fashion to generate the final set of pseudo-labels that are used to recognize out-of-distribution objects (Section~\ref{sec: boundless}). The final output consists of the segmentation map of an image in the target domain.}
    \label{fig: overview}
     
\end{figure*}

\subsection{Open-Set and Boundless DA}

%In most datasets, the class labels are not uniform. 
If the set of labels in the source is equal to the set of labels in the target, then this type of DA is known as \textit{closed-set DA}. On the other hand, if the target domain contains additional class labels that are not present in the source domain, then this type of DA is called \textit{open-set DA}. In open-set DA, the additional class labels in the target domain that do not belong to the source domain are labeled as an ``unknown'' class~\cite{toldo2020unsupervisedreview}. While open-set DA has been proposed for object detection and classification~\cite{openset_original,osbp,sta}, 
they don't extend well for pixel-level tasks like semantic segmentation. 

An extension to open-set DA is boundless DA, where the extra classes present in the target domain are explicitly labeled. Boundless DA has been recently studied by~\cite{bucher2020buda} for semantic segmentation, where the authors successfully classify open-set classes, but at the cost of degraded accuracy on the closed-set categories.

\subsection{Multi-Source Domain Adaptation}

%All the methods described above use a single dataset as the source domain. To leverage all of the available data through multiple datasets, single-source methods combine the data from multiple datasets into one source and proceed as usual. However, empirical studies have shown evidence that this approach often results in a lower performance~\cite{zhao2020multireview}. Therefore, improved methods for multi-source DA are needed.

While multi-source DA has been extensively studied in the context of other perception tasks like object recognition and classification~\cite{guo2020multi,lin2020multi,zhao2019multi,wang2020learningmulti,yang2020curriculummulti}, 
it has not been explored in detail for semantic segmentation \cite{zhao2019multisemantic,zhao2018adversarial}. Prior approaches in multi-source DA suffer from heavy overhead in terms of the requirement of data from all sources at every point of training. In contrast, BoMuDA requires only the pre-trained models, along with data from a chosen source `best source' domain. %In fact, the first approach for multi-source domain adaptive semantic segmentation was only recently proposed by ~\cite{zhao2019multisemantic}. The authors propose a ``sim2real'' technique where they adapt simulation-based source domains to real-world target domains. How

We present the first method for unsupervised multi-source boundless domain adaptive semantic segmentation (See Figure~\ref{fig: venn_diagram}). However, our approach can be generally applied towards domain adaptation in different perception tasks such as object recognition. This is a part of our future work.

\section{BoMuDANet}
In this section, we formally specify our problem, introduce the notation and present details of our neural network used for visual scene understanding in unstructured traffic environments.

\subsection{Problem Setup and Notation}
\label{sec: problem_setup}

% We consider the problem of semantic segmentation in unconstrained traffic videos in the unsupervised multi-source boundless domain adaptation training regime.

Given an RGB image or video of unconstrained traffic selected from the target domain at test time, our goal is to identify the correct object class label of each pixel. In the training phase, we are provided with a set $\mathcal{S}$ of $N$ source domains, in which each source domain is represented as $S_i$, where $i = 1, 2, \ldots, N$, and one target domain $T$.  The set of all categories in the target domain is denoted by $\mathcal{C}_T$, while the set of all categories in the $i^\textrm{th}$ source domain is denoted by $\mathcal{C}_i$. In the boundless DA setting, the target domain may consist of open-set categories \textit{i.e.} classes that are not present in any of the source domains. More formally, $\mathcal{C}_T \setminus \{ \cup_i \mathcal{C}_i \} \neq \emptyset$. 

The output probability map\footnote{Each value in this map is the probability of the pixel, in the corresponding location in the input image, belonging to class $\mc{C}_i$.} for an input image belonging to the $i^\textrm{th}$ source domain is denoted as $P_i \in \mathbb{R}^{\vts{\mc{C}_i} \times h \times w}$, while the ground truth label for the same image is denoted by $y_i \in \mathbb{R}^{h \times w}$. In the unsupervised DA setting, the ground-truth labels for target domain images are not available. We present BoMuDANet in Section ~\ref{sec: multi}.

% \section{Method}
% \label{sec: method}

\subsection{Overview}
\label{sec: multi}

Our method, BoMuDANet (Figure~\ref{fig: overview}), trains a deep neural network using a novel self-training algorithm that we call the Alternating-Incremental (Alt-Inc) algorithm (Section~\ref{subsec: multi-source_DA}). The input consists of images and their corresponding labels from multiple source domains and images from the target domain. The Alt-Inc algorithm generates probability maps corresponding to the target domain image.

% The probability predictions of the alternating-incremental algorithm are used for generating the final output in which the open-set classes are explicitly recognised, without sacrificing performance on the shared classes. 

\subsection{Alternating-Incremental Algorithm for Multi-Source DA}
\label{subsec: multi-source_DA}
The main challenge with unsupervised multi-source domain adaptation is in setting up a cost function \cite{zhao2020multireview} for training the deep neural network (DNN). This is due to two reasons: $(i)$ absence of target domain labels and $(ii)$ variations between different source domains, and each source and target domain. In the proposed approach, we use the idea of ``pseudo'' labels to act in place of the missing target domain labels. The pseudo labels are the class predictions from the `best source' single-source DA model, which is explained below in ``Initialization''. These pseudo labels, along with the pre-trained single-source DA models from the remaining sources are used for training the deep neural network with an improvised cost function. 

In concurrence with the notion of self-training, we observed that repeated re-training of the deep neural network with an enriched cost function each time leads to a significant boost in accuracy. This is because the model weights are optimized after each round of training, which in turn optimizes the pseudo labels, leading to an increasingly accurate cost function to be used in the next round of training\footnote{We provide evidence of the benefits of repeated re-training in the supplementary material.}. As these two optimizations occur in an alternating manner along with domain adaptation from best source and multi-source distillation, incrementally improving the accuracy, we call this training routine (a variation of self-training) the ``Alternating-Incremental'' (Alt-Inc) algorithm.

The motivation behind the Alternating-Incremental algorithm comes from the Expectation-Maximization (EM) algorithm \cite{moon1996expectation}, a classical unsupervised learning algorithm in statistical pattern recognition. The EM algorithm consists of two alternating steps$-$ the E step and the M step. The E step sets up a cost function from observed data, while the M step finds the model parameters that minimize the cost function. Alt-Inc is designed to mimic the principles behind the EM algorithm.

To setup the cost function for the first iteration, the alternating-incremental algorithm selects the best-performing source from the $N$ source domains to generates pseudo-labels that serve as a proxy for the missing target domain labels. This best-performing source dataset is termed as the ``Best-Source''. The inputs to the neural network are images and GT from the ``Best-Source'', pre-trained single-source DA models for all the sources, images from the target domain and the pseudo labels. In summary, the alternating-incremental algorithm (Figure~\ref{fig: overview}) trains the network as follows:

\begin{enumerate}
    \item \textit{Initialize} $\gets$ Best-Source model.
    
    \item \textit{Perform the following in an alternating manner:} \begin{itemize}
        \item Use pseudo labels from previous round of training to set up a cost function for BoMuDANet.
        \item Use this cost function along with the remaining $N$\textminus$1$ pre-trained single-source DA models to train BoMuDANet in an end-to-end manner till convergence.
        % \item The cost function consists of a domain adaptation loss formulation to adapt from the best-source, a knowledge distillation loss formulation to distil information from the remaining sources, and a self-training step that utilizes the pseudo labels. 
    \end{itemize}
    
\end{enumerate}
We now describe each step in detail. 
\subsubsection{Initializing the Best Source Model}
\label{subsec: step0}
We begin by training single-source DA models using each source dataset, and the target dataset, using an adversarial DA paradigm \cite{tsai2018learning}. The single-source domain discriminators (binary classifiers, see architecture below) characterize how indistinguishable the target domain is from each source domain. The output of the discriminators averaged over all target images characterizes the dissimilarity between each source domain and the target domain. The source domain with the least dissimilarity is selected as the ``best source''. The deep neural network (DNN) used to train the best source-target pair is termed the ``Best-Source'' model.

%The source dataset that is closest Let $\textrm{mAcc}_i$ denote the mean accuracy of the $i^\textrm{th}$ source domain. % (computed using Equation~\ref{eq: macc}). 
%Then, the source with the highest mean accuracy is selected as the ``best source'',

%\[ S_\textrm{bs} = \argmax_{S_i \in \mathcal{S}} \textrm{mAcc}_i. \]

% denote the output probability maps of each of these networks. 

\label{subsec: step1}
%Our goal is to generate, and train, pseudo labels that can be used to set up an approximated cost function for self-training the Enriched Best-Source model (Figure~\ref{fig: alt-inc-steps}).

\paragraph{Architecture:} Consistent with adversarial domain adaptation \cite{tsai2018learning}, our network consists of a DNN for semantic segmentation, and domain discriminators. The backbone of the DNN consists of SOTA architectures such as VGG-16 \cite{simonyan2014very}, 
Dilated Residual Network \cite{yu2017dilated}, or DeepLab \cite{chen2017deeplab} (we experiment with different backbones in Section \ref{sec: experiments}). Domain discriminators are neural networks that aim to distinguish whether the predicted segmentation map is from the source or target.

\paragraph{Training:} The inputs to this model consist of raw images and GT from the best source domain, raw images from the target domain $T$, and the pseudo labels. The model weights are initialized with parameters corresponding to the Best-Source baseline obtained in the initialization step. The cost function consists of a domain adaptation loss formulation to adapt from the best-source, a knowledge distillation loss formulation to selectively distil relevant  information from the remaining sources, and a self-training step that utilizes the pseudo labels. We now describe the three loss functions that are used to train the network:

\begin{itemize}
    \item \textbf{The supervised loss function $(\mc{L}_\textrm{sup})$}: This is the standard cross entropy supervised loss function that is used to minimize the distance between the probability map outputs and the ground truth labels.
    \begin{equation}
    \mc{L}_\textrm{sup} = - \sum_{h,w} \sum_{c \in \mc{C}_\textrm{bs}} y_{\textrm{bs}}\log(P),%_\textrm{Alt-Inc}),
    \end{equation}
\noindent where $c$ denotes the object category, $h,w$ denote the height and width of the input images, and $P%_\textrm{bs}
\in \mathbb{R}^{\vts{\mc{C}_\textrm{bs}} \times h \times w }$ is the output of the model on source domain images.

\item \textbf{The unsupervised loss function $(\mathcal{L}_\textrm{unsup})$}: For each target image, we use the trained model from the previous iteration of Alt-Inc to generate pseudo labels for self-training. %we use a curriculum manager~\cite{yang2020curriculummulti} to select the closest source, $S_\textrm{cm}$ to which the target image may belong. Note that we do not use any labeled data in the target domain. Let $S_cm$ denote the source model that best represents(in terms of segmentation feature maps) the given target image. The similarity between each source domain and the given target image is determined by the single source DA discriminators.
The pseudo labels are generated using the probability map predictions, $P%_\textrm{Alt-Inc} 
\in \mathbb{R}^{\vts{\mc{C}%_\textrm{Alt-Inc}
} \times h \times w}$. More formally,

\begin{equation}
    y_\textrm{pseudo} = \argmax_{c \in \mc{C}}\textrm{Softmax}(P%_\textrm{Alt-Inc}
    ).
    \label{eq: pseudo}
\end{equation}

The pseudo-label $y_\textrm{pseudo}$ is used in the unsupervised cross entropy loss function, $\mc{L}_\textrm{unsup}$, as follows,

\begin{equation}
     \mc{L}_\textrm{unsup} = - \sum_{h,w} \sum_{c\in \mc{C}%_\textrm{bs}
     } y_\textrm{pseudo}\log(P%_\textrm{bs}
     ).
     \label{eq: loss_unsup}
\end{equation}

\item  \textbf{Multi-source distillation $(\mc{L}_\textrm{distill})$}: From each of the single source DA networks, we generate their corresponding target domain probability maps $P_i, i \in [N]$. To selectively impart relevant knowledge from various sources, the target domain predictions of BoMuDANet are distilled using a weighted combination of KL divergence \cite{liu2019structured} loss terms corresponding to each of the single-source DA predictions. Kl divergence aligns probability distributions and a weighted combination of KL divergence from multiple sources aids in selective extraction of relevant knowledge. The weights $(w_i)$ are determined by the dissimilarity between each source-target pair (see initialization step above). %the performance of the single source DA networks. 

\begin{equation}
        \mc{L}_\textrm{distil} = \sum_{i} w_i \times KL(P_\textrm{bs}||P_i).
        \label{eq: loss_distill}
\end{equation}
    
\end{itemize}

% \begin{figure}[t]
%     \centering
%     % \captionsetup[subfigure]{labelformat=empty}
%     \includegraphics[width = \columnwidth]{Images/step1.png}
%     \label{fig: bdd_base}
%     \caption{\textbf{AltInc: Enriching the Best-Source model (Section~\ref{subsec: step1}):} Here, the best source model, selected during initialization (Section~\ref{subsec: step0}), is trained in a self-training paradigm to generate enriched pseudo-labels. These enriched pseudo-labels are iteratively used to further refine the enriched best-source model.}
%     \label{fig: alt-inc-steps}
%     % \vspace{-10pt}
%     \end{figure}

The three loss functions are combined as follows:
\begin{equation}
    \mc{L}_\textrm{overall} = \lambda_\textrm{sup}\mc{L}_\textrm{sup} + \lambda_\textrm{unsup}\mc{L}_\textrm{unsup} + \lambda_\textrm{distil}\mc{L}_\textrm{distil},
    \label{eq:overalleq_altinc}
    \end{equation}
where $\lambda_\textrm{sup}$, $\lambda_\textrm{unsup}$, $\lambda_\textrm{distil}$ denote the hyperparameters for the respective loss terms. The domain discriminators are trained in an adversarial \cite{tsai2018learning} fashion. 

\subsubsection{AltInc-EM proof}
In this subsection, we mathematically demonstrate that the Alt-Inc self-training algorithm is an expectation maximization algorithm \cite{em_proof}. The expectation maximization works by iteratively setting up an improved cost function and maximizing it by alternating between the Expectation (E) and Maximization (M) steps.\\
\textbf{E step:} In the EM algorithm, the expected value of the log likelihood of the unseen data $Y$ is computed given the parameter estimate $\theta$ and known data $X$. 
\begin{equation*}
    L(X,\theta) = p(X/\theta) = \int_{y} p(X/\theta) dy
\end{equation*}
In Alt-Inc, we use the estimated network parameters to compute pseudo labels (Equation \ref{eq: pseudo}), which is used to set-up the loss function (Equation \ref{eq:overalleq_altinc}). $X$ corresponds to the best source domain data, and pre-trained source models for the other sources. $Y$ corresponds to the target domain, and $\theta$ corresponds to the parameters of the neural network being optimized. This loss function is a combination of KL divergence and cross entropy loss terms, which are variations of log-likelihood.\\
\textbf{M step:} In the EM algorithm, this step corresponds to maximizing the log-likelihood cost function set-up in the E step. In Alt-Inc, we minimize the cost function. Minimizing KL divergence and cross-entropy is mathematically equivalent to maximizing the cost function \cite{ce_em,kl_em}. 

\subsection{Boundless Domain Adaptation}
\label{sec: boundless}

\begin{table*}
% \footnotesize
\centering
% \begin{center}
\resizebox{\textwidth}{!}{
\begin{tabular}{c c c c c c c c c c c c c c c c c c c c c c}
\toprule
Model & Experiment \Bstrut & mIoU ($\uparrow$) & mAcc ($\uparrow$) & \rotatebox{0}{Road} & \rotatebox{0}{SW} & \rotatebox{0}{Bldg} & \rotatebox{0}{Wall} & \rotatebox{0}{Fnc} & \rotatebox{0}{Pole} & \rotatebox{0}{Lt} & \rotatebox{0}{Sign} & \rotatebox{0}{Veg} & \rotatebox{0}{Trn} & \rotatebox{0}{Sky} & \rotatebox{0}{Ped} & \rotatebox{0}{Rdr} & \rotatebox{0}{Car} & \rotatebox{0}{Trk} & \rotatebox{0}{Bus} & \rotatebox{0}{Mb} & \rotatebox{0}{Bike} \\ 
\toprule
\multicolumn{22}{c}{\Tstrut I. CS, BDD, GTA $\longrightarrow$ IDD (Baseline:~\cite{tsai2018learning})} \Bstrut\\
% \multicolumn{21}{c}{AdaptSegNet Baseline \cite{tsai2018learning}, Vanilla GAN}\\
% \midrule
\multirow{3}{*}{Baselines} & CS$\longrightarrow$IDD \Tstrut& 24.43 & 65.23& 82.46&	22.55&	25.93&	13.22&	9.30&	15.26&	1.92&	19.02&	75.16&	20.41&	29.54&	31.37&	\textbf{8.12}&	49.81&	8.53&	10.41& \textbf{10.29}&	6.55\\
&GTA$\longrightarrow$IDD & 26.74 & 75.40 & 79.83&	9.54&	44.12	&\textbf{16.58}&	12.16&	17.59&	0.85&	14.35&	65.36&	18.20&	82.61&	22.90&	6.56&	41.53&	24.13&	15.40& 9.02 &	0.76\\
&BDD$\longrightarrow$IDD & 35.75 & \textbf{85.65} & 93.33 &	27.17 &	59.77 &	13.18	& 15.56 &	\textbf{21.03} &	\textbf{3.65} &	29.93 &	80.52 &	33.21 &	93.64 &	30.62 &	5.59 &	53.03 &	38.34 &	32.24 & 6.46 &	6.27 \\

\multirow{1}{*}{BoMuDANet}& Multi-source & \textbf{37.66}	& \textcolor{blue}{\textbf{86.50}} &	\textcolor{blue}{\textbf{94.02}} &	\textcolor{blue}{\textbf{31.89}}&	\textbf{61.79}&	\textcolor{blue}{\textbf{15.51}}&	\textbf{16.89}&	\textcolor{blue}{\textbf{20.61}}&	2.73&	\textcolor{blue}{\textbf{35.43}}&	\textcolor{blue}{\textbf{81.75}}&	\textcolor{blue}{\textbf{36.52}}&	\textcolor{blue}{\textbf{94.16}}&	\textbf{32.12}&	4.67&	\textcolor{blue}{\textbf{54.74}}&	\textbf{42.64}&	\textcolor{blue}{\textbf{38.61}}&		5.42&	\textcolor{blue}{\textbf{8.51}} \\
\midrule
\multicolumn{22}{c}{\Tstrut II. SC, BDD, GTA $\longrightarrow$ IDD (Baseline:~\cite{tsai2018learning})} \Bstrut\\
% \multicolumn{21}{c}{\textit{(ii) Sources: GTA, Berkeley Deep Drive, Synscapes}}\\
% \multicolumn{21}{c}{\textit{AdaptSegNet Baseline \cite{tsai2018learning}, Vanilla GAN}}\\

% \midrule
\multirow{3}{*}{Baselines} & SC $\longrightarrow$IDD \Tstrut& 31.55 & 83.04 & 92.46	& 21.25&	52.59&	4.61&	7.87&	17.02&	2.73&	12.60&	77.52&	4.43&	92.38&	31.54&	\textbf{23.32}&	\textbf{66.59}&	4.09&	18.35&		\textbf{27.27}&	\textbf{11.25}\\
%\multirow{2}{*}{Alt-Inc} 

&GTA$\longrightarrow$IDD & 26.74 & 75.40 & 79.83&	9.54&	44.12	&\textbf{16.58}&	12.16&	17.59&	0.85&	14.35&	65.36&	18.20&	82.61&	22.90&	6.56&	41.53&	24.13&	15.40& 9.02 &	0.76\\
&BDD$\longrightarrow$IDD & 35.75 & 85.65 & 93.33 &	27.17 &	59.77 &	13.18	& 15.56 &	21.03 &	\textbf{3.65} &	29.93 &	80.52 &	33.21 &	93.64 &	30.62 &	5.59 &	53.03 &	38.34 &	32.24 & 6.46 &	6.27 \\

BoMuDANet & Multi-source  & \textbf{36.93} & \textbf{86.30} & \textbf{93.82} &	\textbf{30.53}&	\textbf{61.13}&	\textcolor{blue}{\textbf{13.34}}&	\textbf{16.43}&	\textbf{21.21}&	\textcolor{blue}{\textbf{3.57}}&	\textbf{34.90}&	\textbf{81.64}&	\textbf{34.54}&	\textbf{94.19}&	\textbf{31.70}&	4.64&		\textcolor{blue}{\textbf{53.48}}&	\textbf{40.77}&	\textbf{35.54}&		5.68&		\textcolor{blue}{\textbf{7.64}}\\
\midrule
\multicolumn{22}{c}{\Tstrut III. CS, BDD, GTA $\longrightarrow$ IDD (Baseline:~\cite{vu2019advent})} \Bstrut\\

% \midrule
\multirow{3}{*}{Baselines}&CS $\longrightarrow$ IDD \Tstrut & 38.53& 86.68&	93.67&	27.08&	\textbf{64.62}&\textbf{	25.89}&	17.80&	23.39&	4.18&	31.29&	\textbf{83.06}&	29.83&	94.22&	32.28&	11.18&	61.68&	39.86&	33.32&		12.08&	8.23	 \\
&GTA $\longrightarrow$ IDD & 35.85 & 84.64 & 89.96 & 14.06 & 61.14 & 22.24 & 20.10 & 19.17 & 4.34 & 19.88 & 77.15 & 28.84 & 92.14 & 27.03 & \textbf{11.98} & \textbf{62.87} & 41.04 & 34.67 & 13.10 & 5.74 \\
&BDD $\longrightarrow$ IDD & 38.29&	86.74&	\textbf{93.80}&	\textbf{33.33}&	62.57&	14.94&	15.35&	\textbf{23.66}&	3.80&\textbf{	31.95}&	81.72&	34.47&	\textbf{94.26}&	\textbf{33.00}&	8.71&	57.11&	42.87&	39.16&		9.41&	9.22 \\
% Alt-Inc & Multi-source & fill & fill \Bstrut\\
%\multirow{2}{*}{Alt-Inc}
BoMuDANet & Multi-source & \textbf{39.23} &	\textbf{87.18}&	93.18&	\textcolor{blue}{\textbf{29.97}}&	\textcolor{blue}{\textbf{63.46}}&	\textcolor{blue}{\textbf{24.18}}&	\textbf{20.97}&	\textcolor{blue}{\textbf{19.18}}&	\textbf{4.56}&	25.64&	\textcolor{blue}{\textbf{81.99}}&\textbf{35.39}&	94.19&	30.06&\textcolor{blue}{\textbf{	11.23}}&	\textcolor{blue}{\textbf{62.01}}&	\textbf{46.65}&	\textbf{39.30} & \textbf{13.39}&	\textbf{10.87} \\

\bottomrule

\end{tabular}
}
\caption{\textbf{Main Results: }We evaluate BoMuDANet on IDD using CityScapes (CS), Berkeley Deep Drive (BDD), SynScapes (SC), and GTA as sources. Higher ($\uparrow$) mIoU and mAcc indicates direction of better performance. \textbf{Bold} indicates best while \textcolor{blue}{\textbf{blue}} indicates second-best. Experiments I and II differ with respect to the sources, while experiment III differs with respect to the baseline used. \textbf{Conclusion:} Our unsupervised multi-source Alt-Inc algorithm outperforms the single-source baselines by $3.3 \% - 54.15\%$.}

\label{tab:idd_mainresults}
% \vspace{-10pt}
% \end{center}
\end{table*}

We present a new method for performing Boundless DA \textit{i.e.} to label categories that exist in the target dataset, but not in any of the source datasets (``open-set'' or ``private'' or ``unknown'' categories). Categories that are common to both the source and the target domains are called ``closed-set'' or ``shared'' or ``known'' categories. The key assumption in our solution is that the open-set categories are physically similar to the closed-set categories. For instance, open-set categories such as auto-rickshaws are similar to vehicles like cars and vans. CityScapes \cite{cs} provides a definition for grouping semantically similar classes in autonomous driving environments. Classes that belong to the same high-level category will have feature maps that are semantically similar, and vice versa. This assumption is mild and is commonly made in many zero-shot learning strategies~\cite{bucher2019zero}. 

The underlying idea behind training our approach on open-set classes is to generate the corresponding pseudo-labels from the labels of the physically similar closed-set categories. More formally, let $y_\textrm{Alt-Inc} \in \mathbb{R}^{h\times w}$ be the final labels obtained using Equation~\ref{eq: pseudo} from the Alt-Inc algorithm. Further, let $o \in \mc{O}$ denote an open-set class from the set of open-set classes, $\mc{O}$, and $\mc{C}_o$ denote the set of closed-set classes that are physically similar to $o$. We apply thresholding on $y_\textrm{Alt-Inc}$ such that pixels with softmax scores lower than a threshold $\tau$ for a physically-similar closed-set class are re-labeled as the open-set class. More formally, let $\hat y_\textrm{pseudo}$ denote the labels after thresholding, then $\hat y_\textrm{pseudo}$ is computed using,

\begin{equation}
    \hat y_\textrm{pseudo} = \mc{T}y_\textrm{Alt-Inc}
    \label{eq: pseuso_open}
\end{equation}

\noindent where $\mc{T}(\cdot)$ is a pixel-level thresholding operator. If $l_{ab}$ denotes the class label of a pixel in the $a^\textrm{th}$ row and $b^\textrm{th}$ column with confidence score $c_{ab}$ (maximum probability value over all classes, as determined by the output probability map of Alt-Inc algorithm), then the threshold operator at $(a,b)$ is defined as,

\[ \mc{T}(a,b) = \begin{cases} 
      l_{ab} \gets o & c_{ab} \leq \tau \ \textrm{and} \ l_{ab} \in \mc{C}_o\\
      l_{ab} & \textrm{otherwise} \\
   \end{cases}
\]

\noindent An alternative to the thresholding operator is to use the KL divergence metric \cite{liu2019structured} to measure the similarity between open-set and closed-set object classes. We empirically observe, however, via an ablation study that thresholding in fact outperforms using the KL divergence metric (See Table~\ref{tab:open_set}). 

%\begin{equation}
%     \hat{\mc{L}}_\textrm{unsup} = - \sum_{h,w} \sum_{c\in %\mc{C}_\textrm{bs} \cup \mc{C}_T} \hat %y_\textrm{pseudo}\log(P_\textrm{bs}).
%     \label{eq: loss_unsup_boundless}
%\end{equation}

%Since the thresholding operator in Equation~\ref{eq: pseuso_open} applies to all pixels in $y_\textrm{pseudo}$, this can produce false positives. To mitigate this issue, we add a training step wherein closed-set predictions determined from the Alt-Inc algorithm are used as the input to a shallow CNN. 

\section{Experiments and Results}
\label{sec: experiments}

We will make all code publicly available. We defer the technical implementation details of the training routine including hyper-parameter selection to the supplementary material. 

%\subsection{Datasets}
\subsection{Datasets and Evaluation Protocol}
\label{subsec: datasets}

%We use four source domain datasets- GTA5~\cite{gta}, SynScapes (SC)~\cite{wrenninge2018synscapes}, CityScapes (CS)~\cite{cs}, and Berkeley Deep Drive (BDD)~\cite{yu2020bdd100k}. 
We use five datasets - GTA5~\cite{gta}, SynScapes (SC)~\cite{wrenninge2018synscapes}, CityScapes (CS)~\cite{cs}, India Driving Dataset (IDD)~\cite{varma2019idd} and Berkeley Deep Drive (BDD)~\cite{yu2020bdd100k}. GTA5 and SC contain synthetic simulated traffic videos while CS and BDD consist of real-world traffic in Europe and the USA, respectively. %Compared to the aforementioned datasets, BDD is challenging and diverse since it contains night scenes, and images captured in adverse weather conditions. Images in the India Driving Dataset (IDD) have been captured in India, during the daytime. 
IDD consists of dense and unconstrained traffic conditions and heterogeneous road agents (e.g. autorickshaws) unobserved in any of the source domains. In addition to containing new objects, the pixel count (per class) in IDD is $5$\textminus$10 \times$ that of CS 

We evaluate our models on the validation set images of the target domain, using the standard segmentation metrics \cite{long2015fully}: Intersection over Union (IoU) and pixel-wise accuracy.

\begin{figure}[t]
    \centering
    \includegraphics[width=\columnwidth, height=3.8cm]{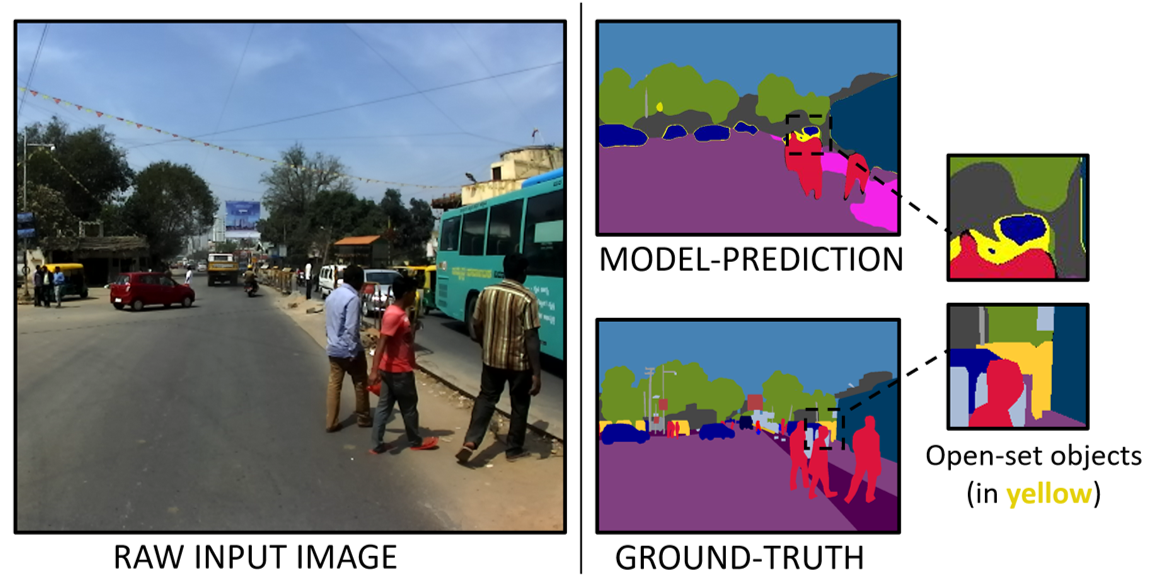}
    \caption{In this example, we demonstrate the benefits of BoMuDANet on an image from the IDD dataset, depicting a mixture of challenging driving conditions. The top image in the second column shows the prediction of our model, and the bottom image shows the ground-truth. We observe that BoMuDANet accurately segments the autorickshaws (open-set object, a new type of vehicle - the third column zooms into the region containing the autorickshaw in the prediction and ground-truth), in addition to handling dense traffic and dirt roads. 
    }
    \label{fig: bomuda_auto}
    
\end{figure}

\begin{table}[t]
% \footnotesize
\centering
% \begin{center}
\resizebox{\columnwidth}{!}{
\begin{tabular}{c c c c c c}
\toprule
 Experiment & mIoU ($\uparrow$) & Car & Truck & Bus & \textbf{Auto} \Bstrut \\
\toprule
\multicolumn{6}{c}{CS, BDD, GTA $\longrightarrow$ IDD, Baseline:~\cite{tsai2018learning}}\\
% \textit{AdaptSegNet Baseline \cite{tsai2018learning}, Vanilla GAN}&&&&&\\
\midrule
Pseudo labeling (PL) & 35.68 & 51.16 & 33.89 & 28.99 & \textbf{9.39} \\
PL + training & 35.72 & 52.18 & 33.93 & 31.65 & \textbf{9.38}\\
\midrule
\midrule
\multicolumn{6}{c}{SC, BDD, GTA $\longrightarrow$ IDD, Baseline:~\cite{tsai2018learning} }\\

% \textit{(ii) Sources: GTA, Berkeley Deep Drive, Synscapes}&&&&&\\
% \textit{AdaptSegNet Baseline \cite{tsai2018learning}, Vanilla GAN}&&&&&\\
\midrule
% \midrule
Pseudo labeling (PL) & 34.60 & 48.36 & 30.78 & 20.82 & \textbf{9.68} \\
 PL + training & 34.40 & 49.14 & 30.44 & 22.59 & \textbf{9.48}\\

 \midrule
 \midrule
\multicolumn{6}{c}{CS, BDD, GTA $\longrightarrow$ IDD, Baseline:~\cite{vu2019advent}}\\

% \textit{(ii) Sources: GTA, Berkeley Deep Drive, Synscapes}&&&&&\\
% \textit{AdaptSegNet Baseline \cite{tsai2018learning}, Vanilla GAN}&&&&&\\
% \midrule
\midrule
Pseudo labeling (PL) & 37.27 & 58.76 & 36.58 & 22.15 &\textbf{11.78} \\
 PL + training & 37.09 & 58.63 & 36.65 & 24.26 &\textbf{11.85}\\
\midrule
\midrule
\multicolumn{6}{c}{Ablation Experiments on IDD, Baseline:~\cite{tsai2018learning}}\\
\midrule
% \midrule
KL Divergence    & 35.43 & 52.12 & 33.99 & 31.46 & 9.29 \\
\bottomrule
\end{tabular}
}
\caption{\textbf{Pseudo labeling strategy for boundless DA:} We show that pseudo labeling can provide semantic information abuot categories that do not belong to any source domain, for instance, \textbf{auto-rickshaws} (Auto) found in the IDD dataset (in \textbf{bold}). Moreover, pseudo labeling is simple in that the generated proxy labels do no need to be re-trained as there is no marked improvement in mIoU. We also show that thresholding outperforms KL divergence via an ablation study. }
% \textbf{Conclusion:} The accuracy of our model on auto-rickshaws (`unknown class') is comparable (and even better in some cases!) to the accuracy on many `known' classes like motorbike, bike and light, Table  \ref{tab:idd_mainresults}.}

\label{tab:open_set}
\vspace{-10pt}
% \end{center}
\end{table}

%\subsection{Evaluation Protocol}
%\label{subsec: evalution_protocol}

%Following the standard procedure in the literature~\cite{bucher2020buda,iiith,chenli}, we use the following two metrics:

%\begin{enumerate}
%    \item Mean Intersection over Union (mIoU) : The IoU score is defined as the amount of overlap between the ground truth mask ($\mathcal{M}_\textrm{GT}$) and the predicted mask $\mathcal{M}_\textrm{Pred}$ for each class. A mask for a class is a set of pixels that belong to that class category. The overlap can be computed by measuring the ratio of the intersection of the two masks to the union of the two masks, 
 %   \begin{equation}
 %       \textrm{IoU} =  \frac{\mathcal{M}_\textrm{GT}\cap \mathcal{M}_\textrm{Pred}}{\mathcal{M}_\textrm{GT} \cup \mathcal{M}_\textrm{Pred}}.
 %   \end{equation}
    
  %  \noindent The Mean IoU or mIoU is computed as the average of the IoU scores corresponding to the different classes. 

  %  \item Mean accuracy (mAcc): The mean accuracy can be defined as the percentage of pixels that are correctly classified according to the ground-truth labels. That is,
    
  %  \begin{equation}
  %      \textrm{mAcc} =  \frac{\textrm{pixels correctly classified}}{\textrm{total number of ground-truth pixels}} \times 100.
  %      \label{eq: macc}
  %  \end{equation}
    
%\end{enumerate}
\begin{figure*}[t]
    \centering
    % \captionsetup[subfigure]{labelformat=empty}
    \begin{subfigure}[b]{0.22\textwidth}
    \includegraphics[width = 3.5cm, height = 1.9cm]{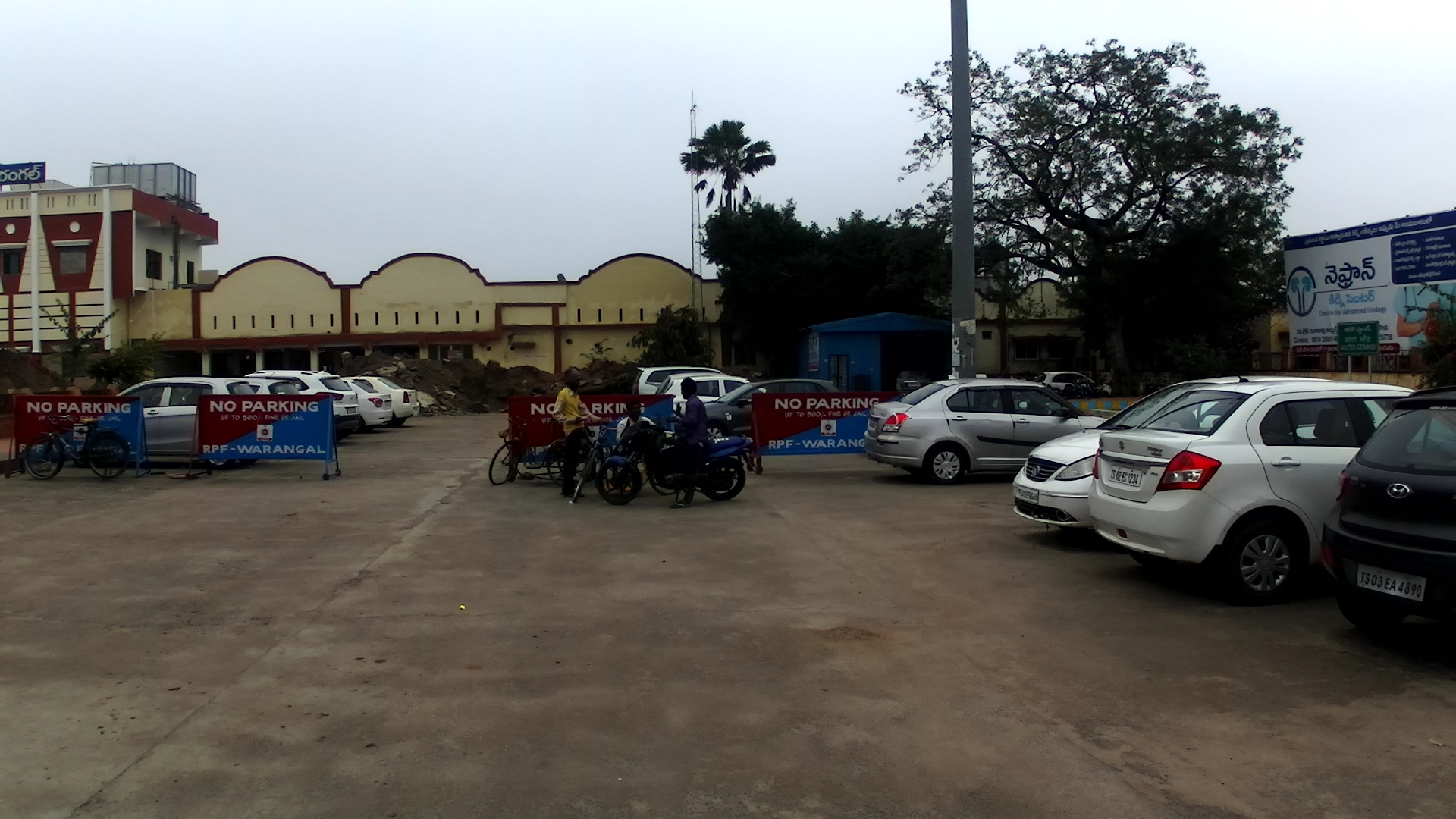}
    \caption{Dirt roads}
    \end{subfigure}
    \begin{subfigure}[b]{0.22\textwidth}
    \includegraphics[width = 3.5cm, height = 1.9cm]{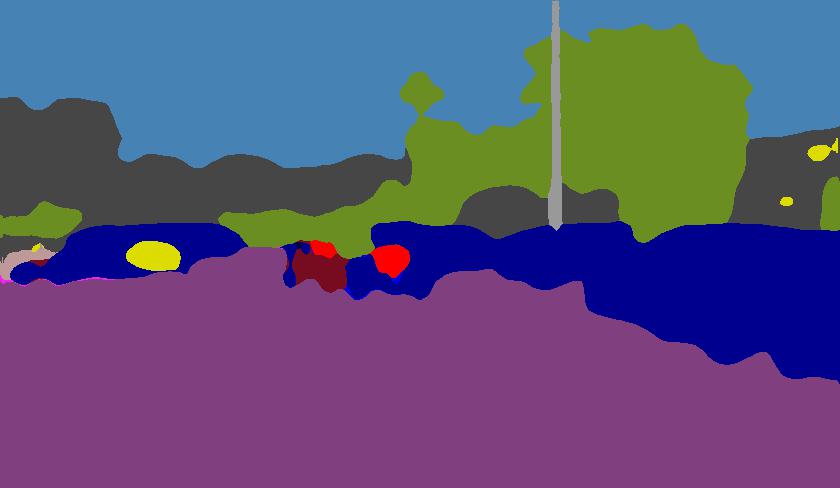}
    \caption{BoMuDANet}
    \end{subfigure}
    \begin{subfigure}[b]{0.22\textwidth}
    \includegraphics[width = 3.5cm, height = 1.9cm]{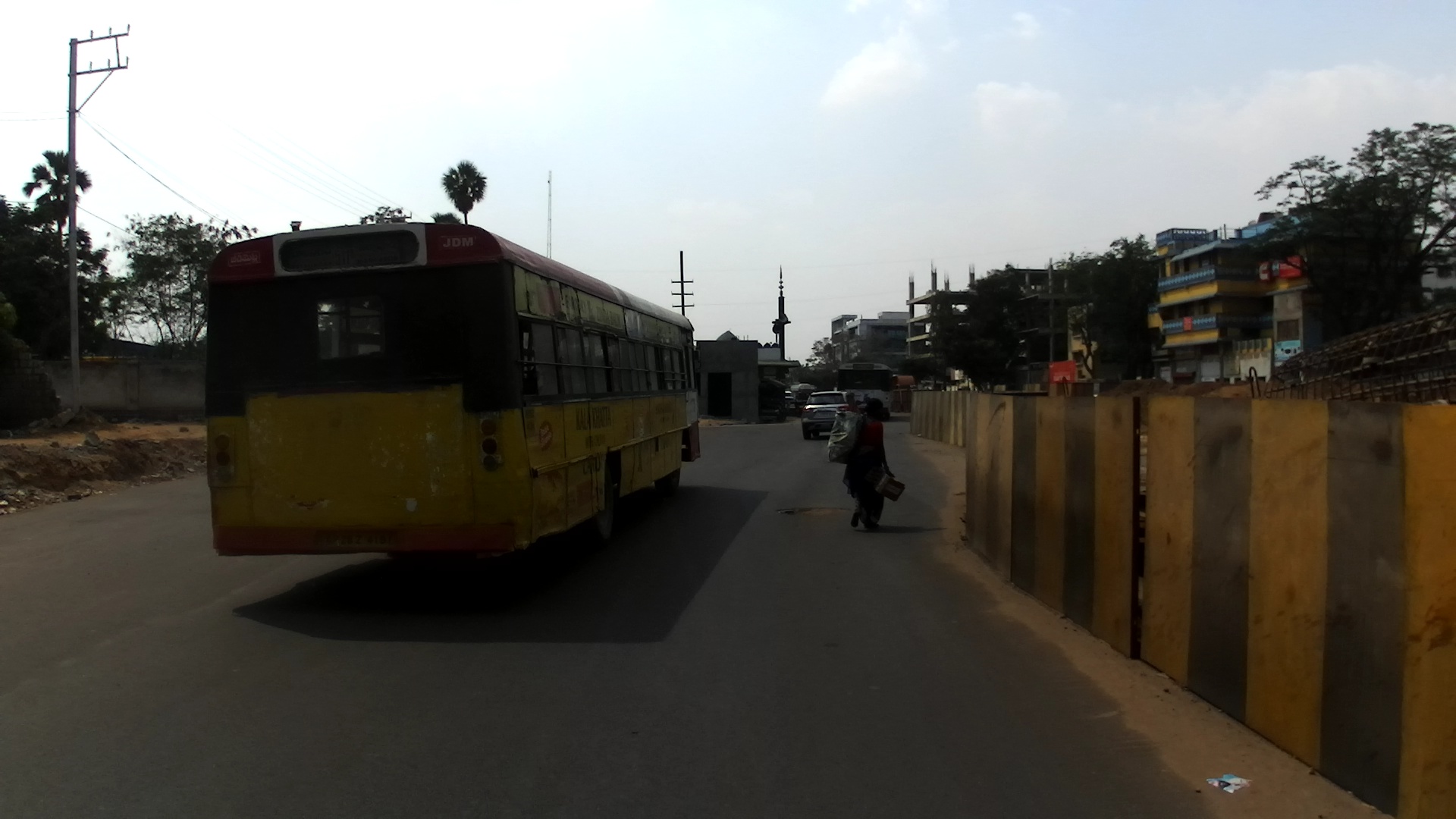}
    \caption{Unmarked lanes}
    \end{subfigure}
    \begin{subfigure}[b]{0.22\textwidth}
    \includegraphics[width = 3.5cm, height = 1.9cm]{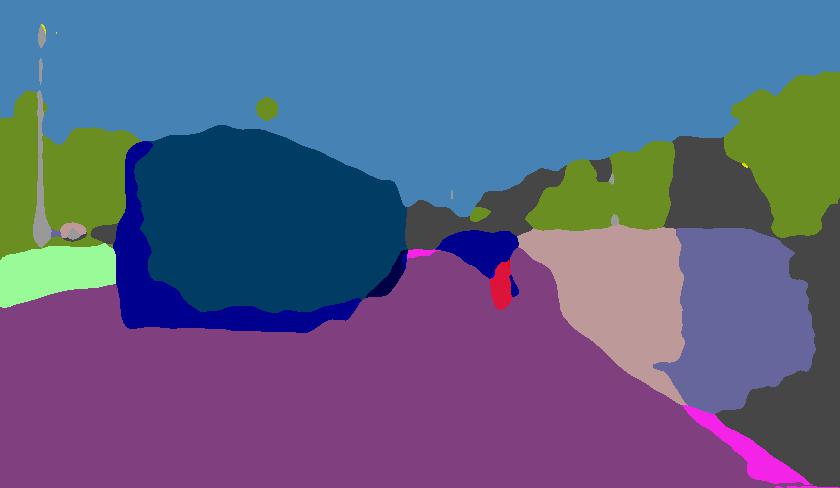}
    \caption{BoMuDANet}
    \end{subfigure}
    \\
    \begin{subfigure}[t]{0.22\textwidth}
    \includegraphics[width = 3.5cm, height = 1.9cm]{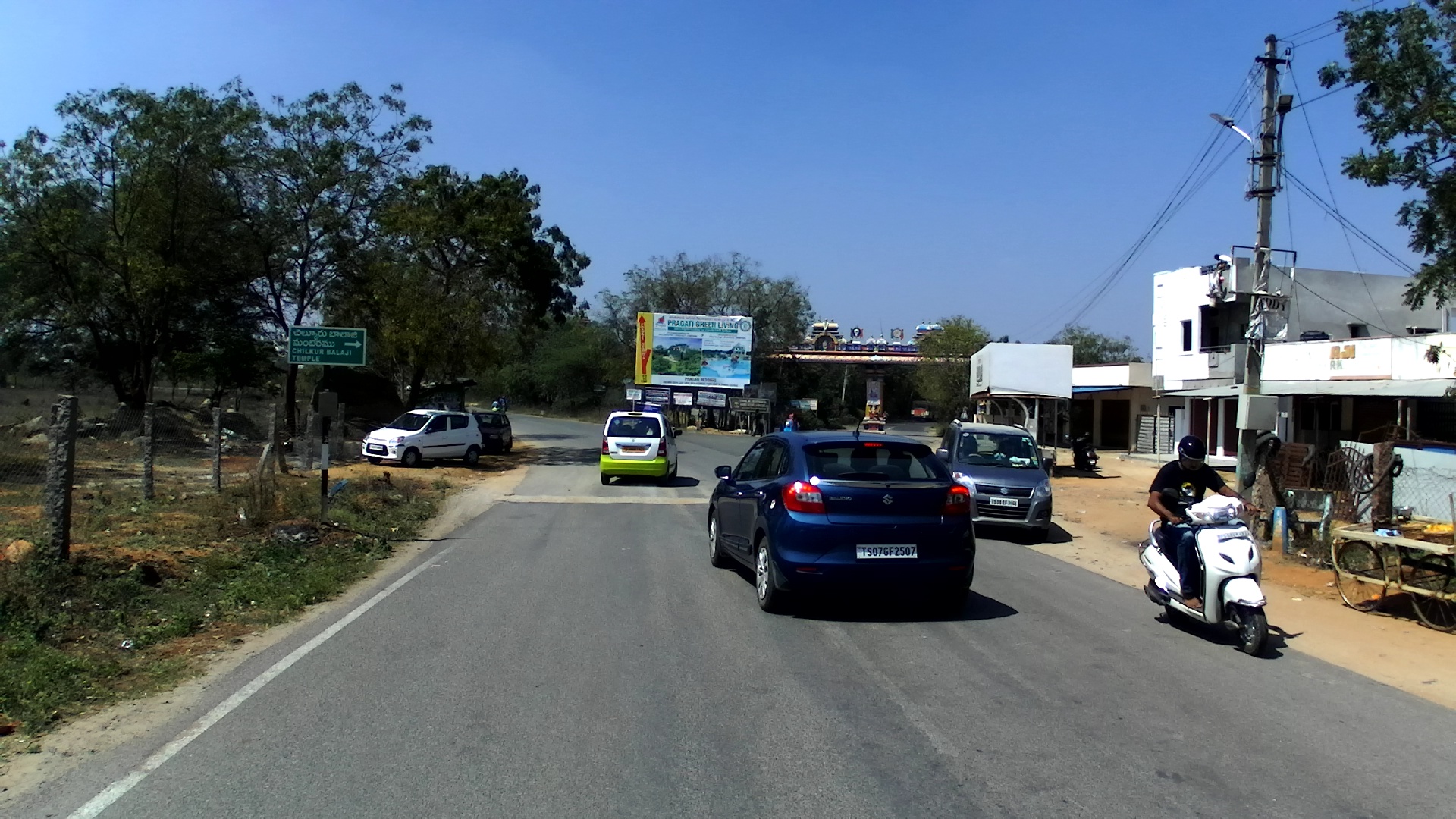}
    \caption{Heavy traffic}
    \end{subfigure}
    \begin{subfigure}[t]{0.22\textwidth}
    \includegraphics[width = 3.5cm, height = 1.9cm]{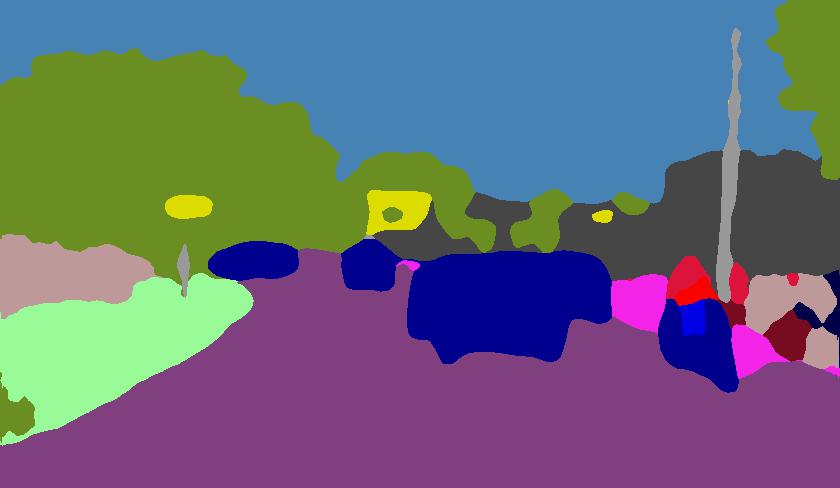}
    \caption{BoMuDANet}
    \end{subfigure}
    \begin{subfigure}[t]{0.22\textwidth}
    \includegraphics[width = 3.5cm, height = 1.9cm]{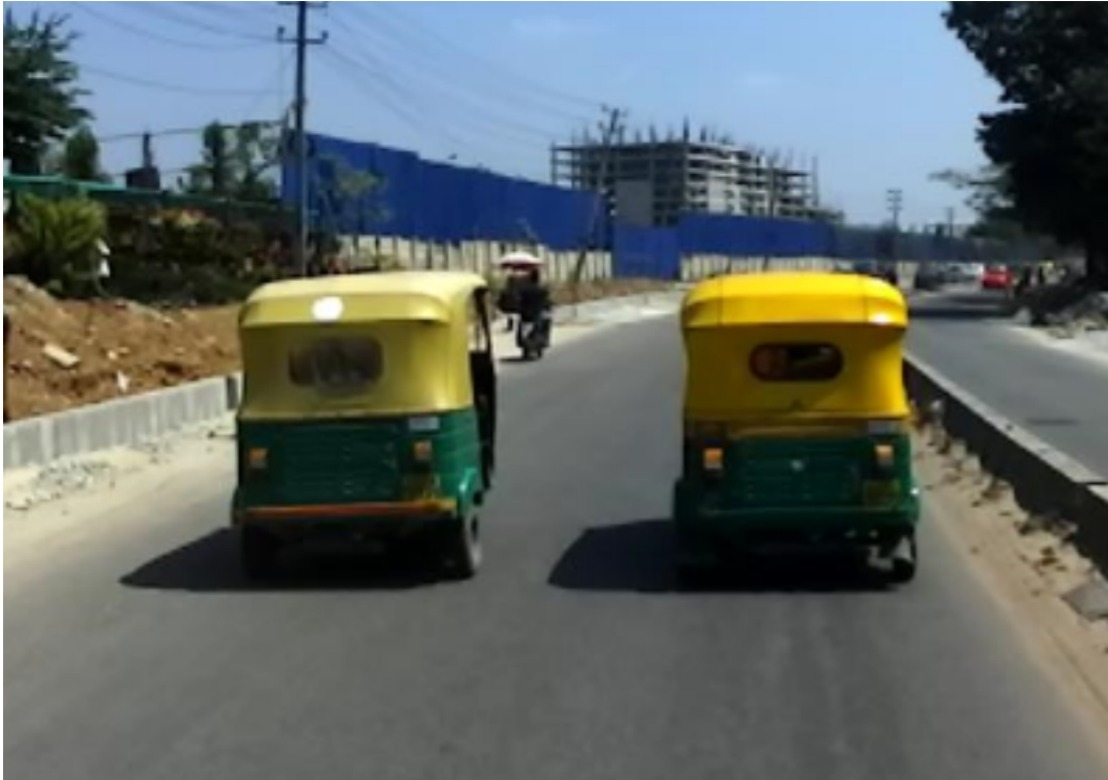}
    \caption{Boundless class}
    \label{fig: boundless_object}
    \end{subfigure}
    \begin{subfigure}[t]{0.22\textwidth}
    \includegraphics[width = 3.5cm, height = 1.9cm]{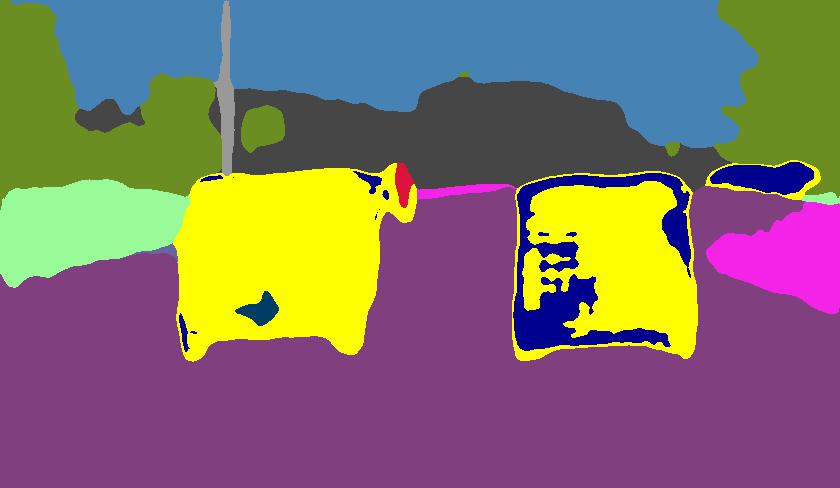}
    \caption{BoMuDANet}
    \label{fig: boundless_object_result}
    \end{subfigure}
    \\
    
    \begin{subfigure}[b]{0.22\textwidth}
    \includegraphics[width = 3.5cm, height = 1.9cm]{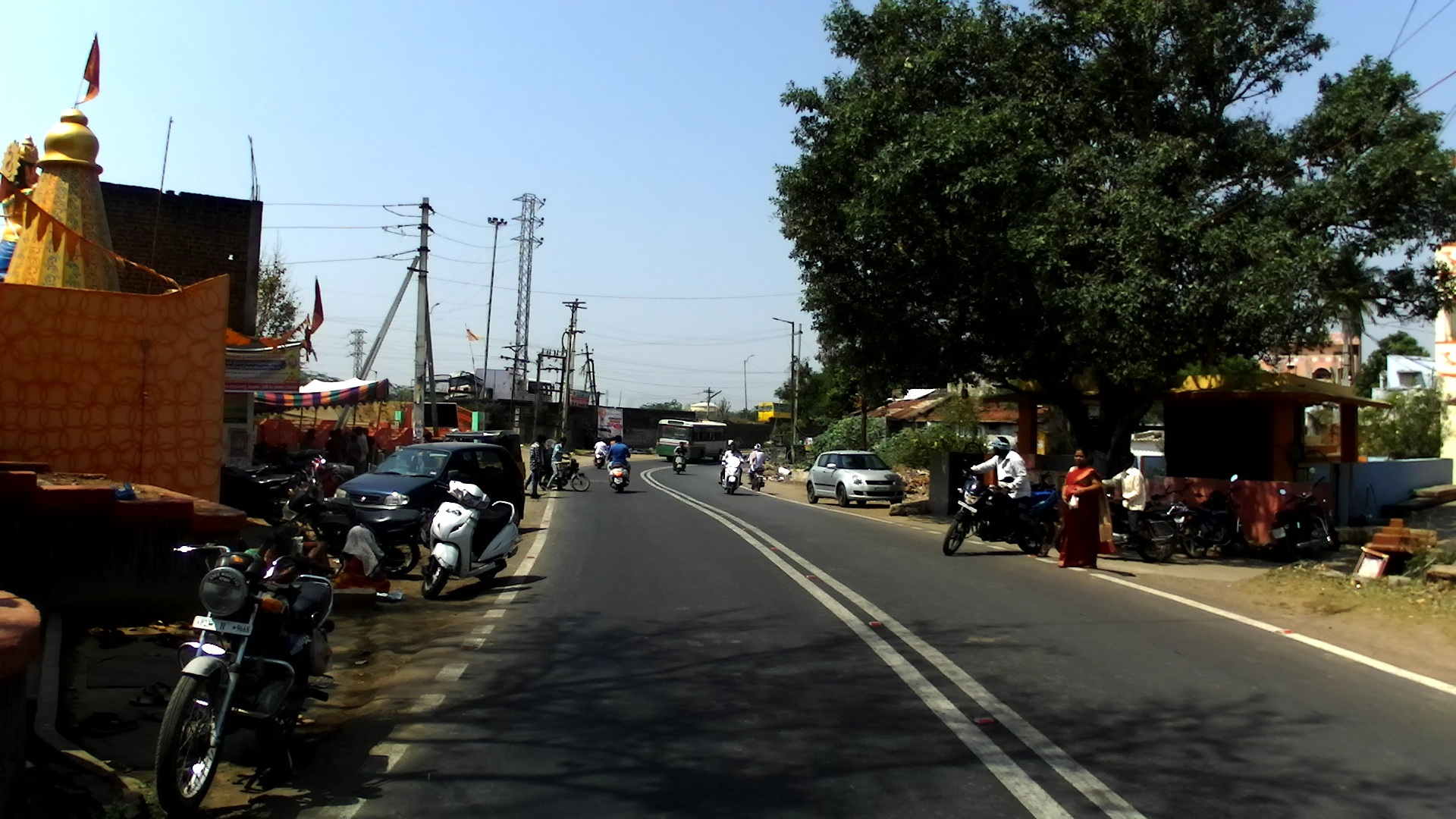}
    \caption{Unstructured roads}
    \end{subfigure}
    \begin{subfigure}[b]{0.22\textwidth}
    \includegraphics[scale=0.15]{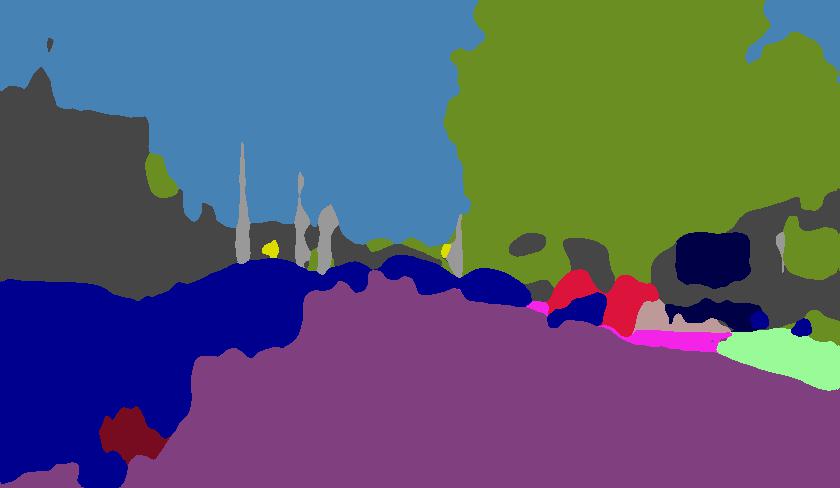}
    \caption{BoMuDANet}
    \end{subfigure}
    \begin{subfigure}[b]{0.22\textwidth}
    \includegraphics[width = 3.5cm, height = 1.9cm]{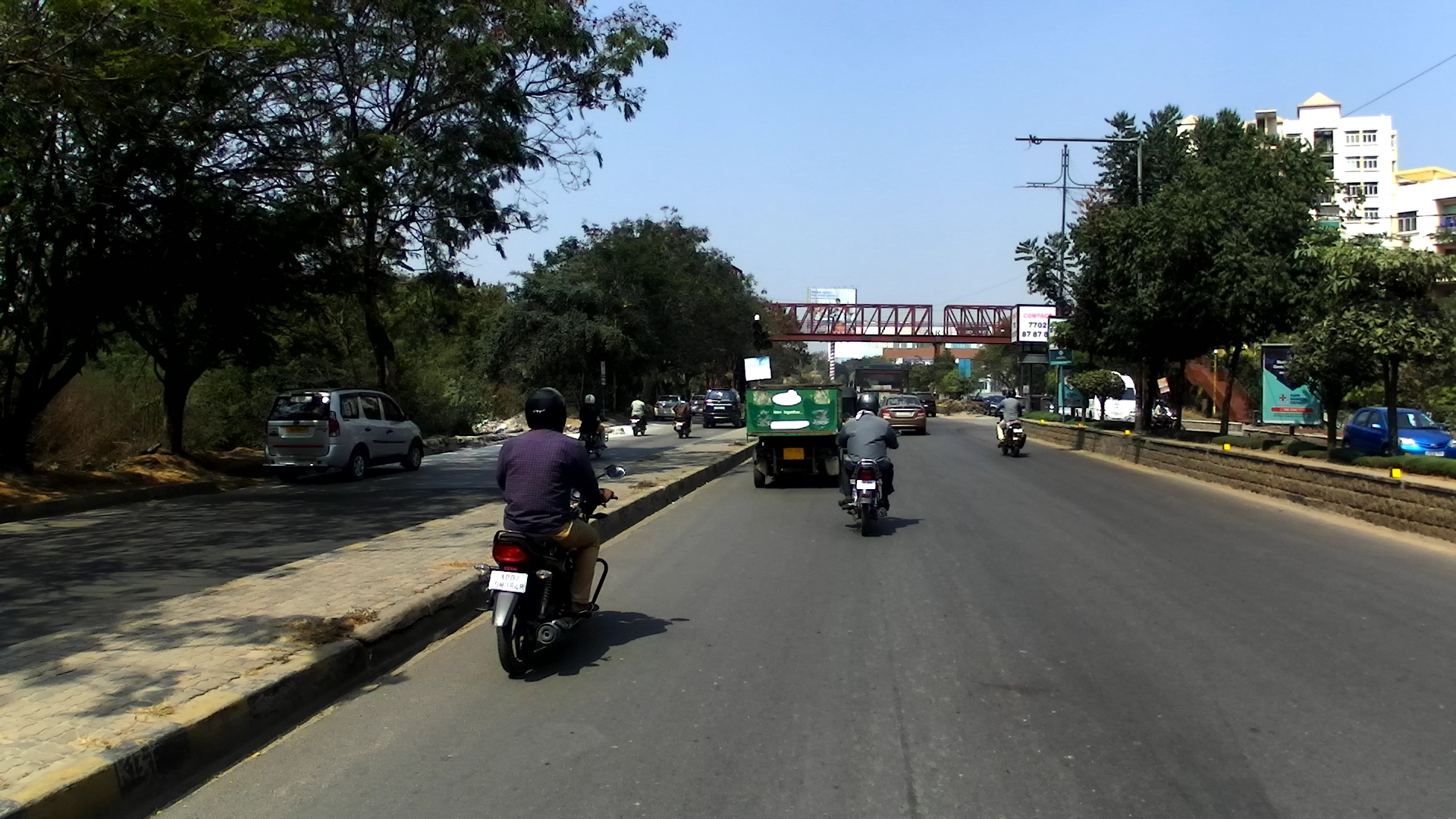}
    \caption{Clear roads}
    \end{subfigure}
    \begin{subfigure}[b]{0.22\textwidth}
    \includegraphics[scale=0.15]{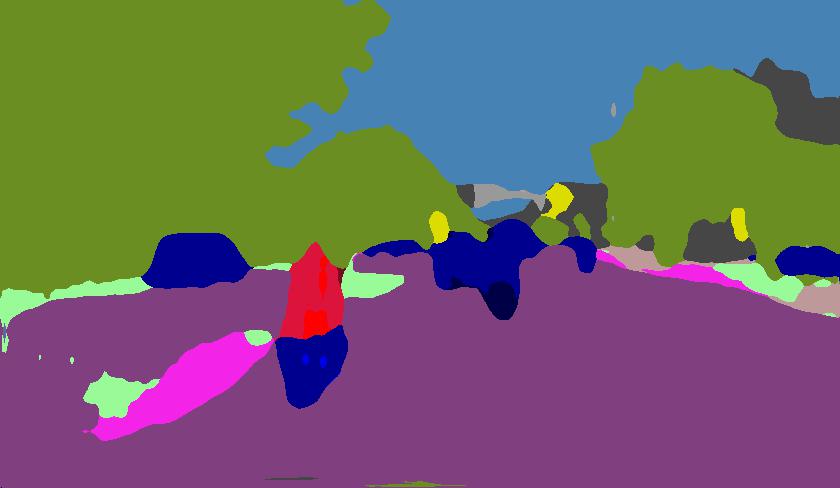}
    \caption{BoMuDANet}
    \end{subfigure}\\
    
     \begin{subfigure}[b]{0.22\textwidth}
    \includegraphics[width = 3.5cm, height = 1.9cm]{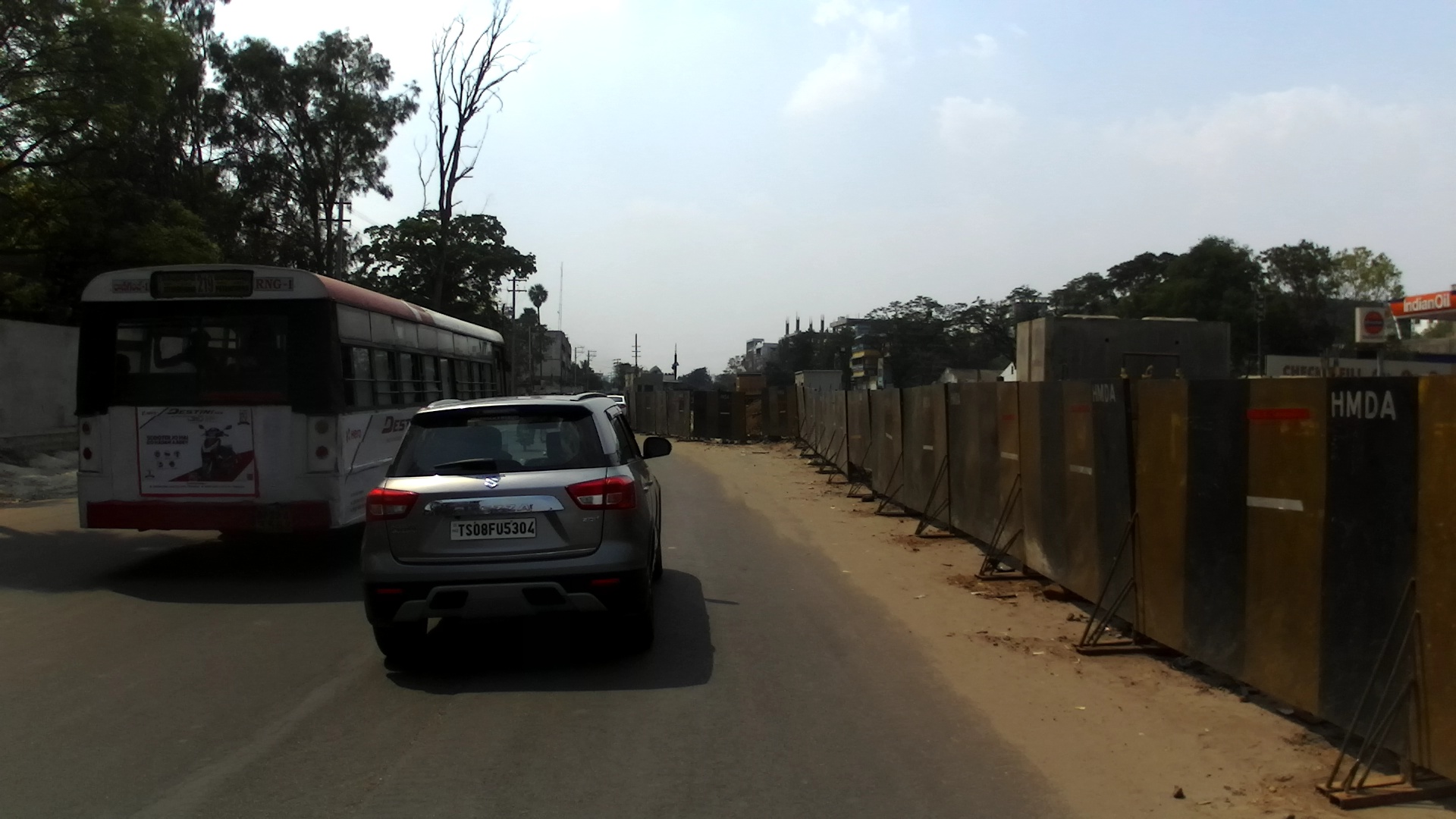}
    \caption{Unstructured roads}
    \end{subfigure}
    \begin{subfigure}[b]{0.22\textwidth}
    \includegraphics[scale=0.15]{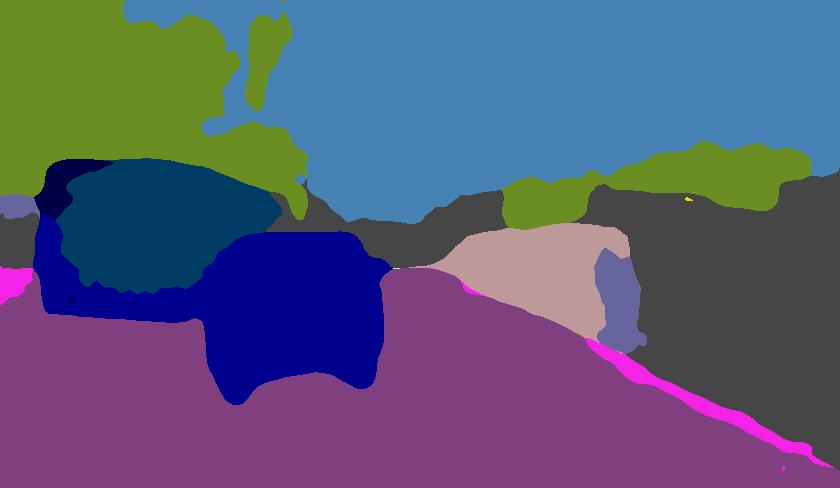}
    \caption{BoMuDANet}
    \end{subfigure}
    \begin{subfigure}[b]{0.22\textwidth}
    \includegraphics[width = 3.5cm, height = 1.9cm]{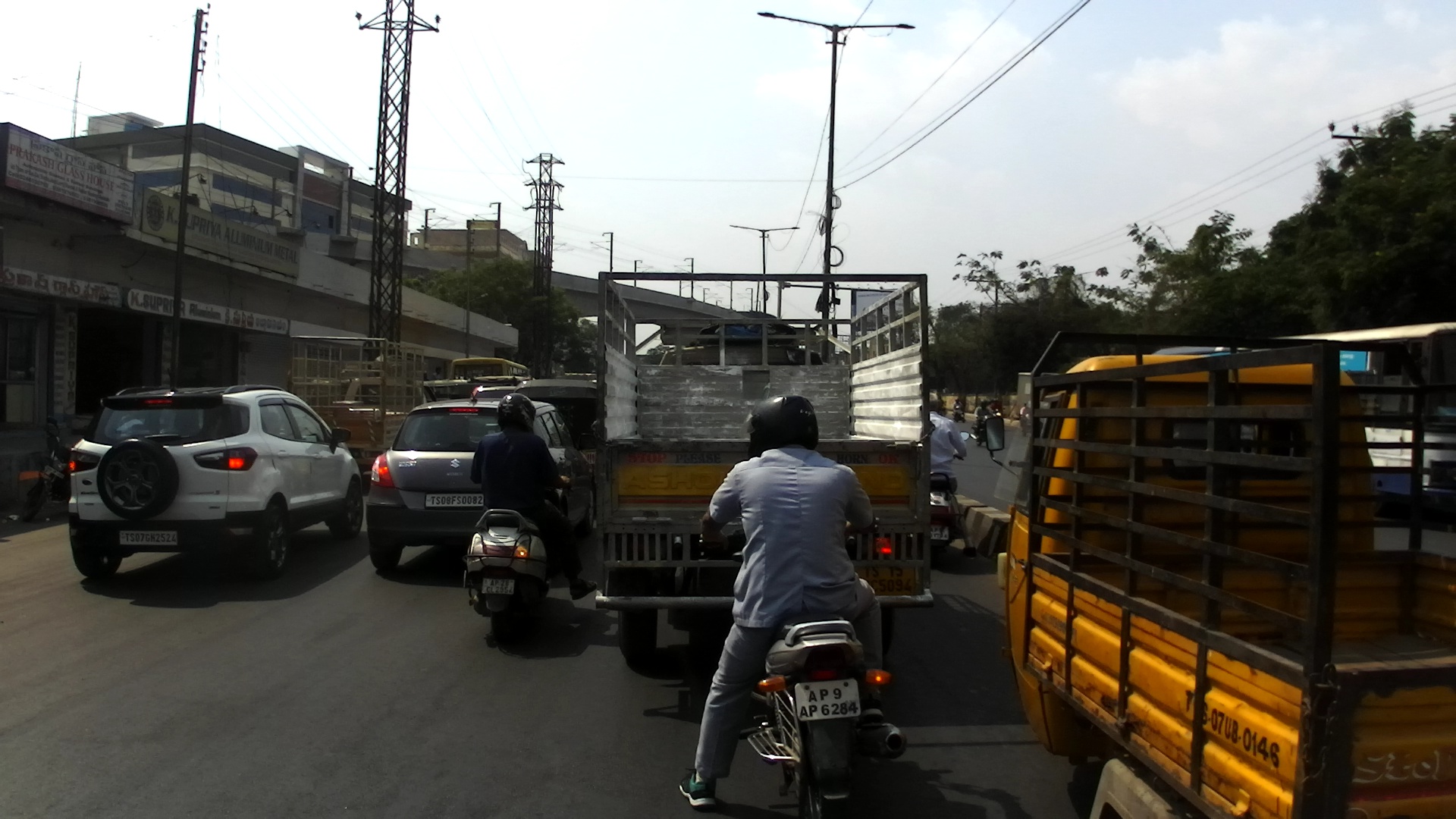}
    \caption{Unstructured and dense traffic}
    \end{subfigure}
    \begin{subfigure}[b]{0.22\textwidth}
    \includegraphics[scale=0.15]{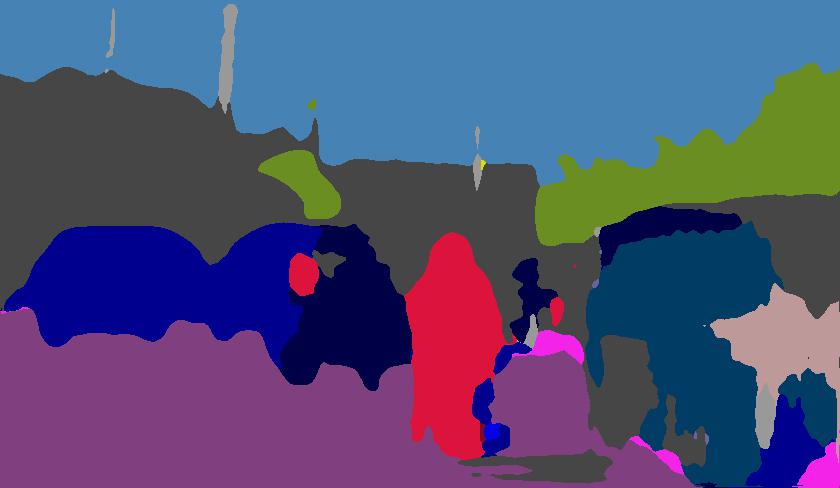}
    \caption{BoMuDANet}
    \end{subfigure}\\
    \begin{subfigure}[b]{0.88\textwidth}
    \includegraphics[scale=0.5]{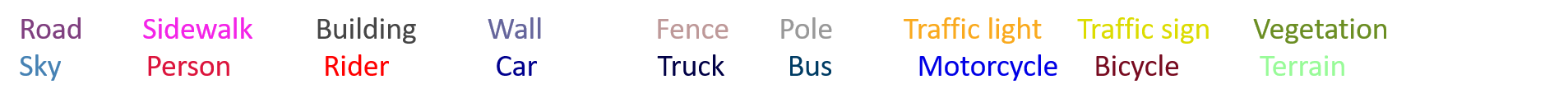}
    %\caption{BoMuDANet}
    \end{subfigure}\\

    \caption{\textbf{Visual Results:} BoMuDANet works well in various unconstrained environments including unmarked lanes, dirt roads, heavy traffic, and boundless category objects (auto-rickshaws) and results in higher accuracy. Each color represents a different object as shown in the color scheme. }
    
    \label{fig:visualisations_altinc}
    % \vspace{-10pt}
\end{figure*}

%\begin{subfigure}[b]{0.22\textwidth}
    %\includegraphics[width = 3.5cm, height = 1.9cm]{Qualitative/frame11395_leftImg8bit.jpg}
    %\caption{Unmarked lanes}
    %\end{subfigure}
    %\begin{subfigure}[b]{0.22\textwidth}
    %\includegraphics[width = 3.5cm, height = 1.9cm]{Qualitative/11395_closedsetpred.jpg}
    %\caption{BoMuDANet}
    %\end{subfigure}

\subsection{Results}

\paragraph{Main Results on IDD (Table~\ref{tab:idd_mainresults}):} We present results of three sets of experiments using IDD as the target dataset in Table~\ref{tab:idd_mainresults}. In each experiment, we compare BoMuDANet with single-source baseline models using the BDD, CS, SC, and GTA datasets, with the BDD dataset selected as the Best-Source. Note that we do not compare with a combination of single source datasets as combining multiple sources and treating them as a single source for DA has been shown to be ineffective~\cite{zhao2020multireview}. We compare with SOTA multi-source DA methods in Section~\ref{subsec: comparison}.

We perform the first experiment with two real datasets (CS, BDD) and one synthetic dataset (GTA5), and show an improvement of $1.91-13.23 (5.34\%-54.15\%)$ mIoU points over the single-source baselines. In the second experiment, we replace the two real source datasets with two synthetic source datasets (SC, GTA) and one real dataset (BDD) with the BDD dataset as the Best Source, and show an improvement of $3.3\%- 38.1\%$ mIoU points over the single-source baselines. By comparing these two sets of experiments, we demonstrate that using multiple real-world datasets is more beneficial than using multiple synthetic datasets.

In the third experiment, we replace the AdaptSegNet \cite{tsai2018learning} backbone with a stronger SOTA backbone ADVENT \cite{vu2019advent}, and use LS GAN for adversarial training instead of Vanilla GAN, and achieve a higher mIoU of 39.23. This suggests that the performance of our approach will increase as newer robust backbone architectures are proposed. We also validate this hypothesis by conducting experiments on structured environments as the target domain.

\paragraph{Results for the Boundless Case:} In Table \ref{tab:open_set}, we show the results for the proposed pseudo labeling strategy for boundless DA method. Note that the thresholding operator, $\tau$, is a tunable hyperparameter. A low value of $\tau$ will create a bias towards the private classes, a high value of $\tau$ will create a bias towards shared classes. A trade-off determines the optimal value of $\tau$ for best performance on both private and shared classes. Typically, tuning between $80\%-90\%$ of max confidence score for the particular class works well.

%Since the thresholding operator in Equation~\ref{eq: pseuso_open} applies to all pixels in $y_\textrm{Alt-Inc}$, it can produce false positives. To mitigate this issue, the unsupervised loss function in Equation~\ref{eq: loss_unsup} can then be used to retrain the network on $\hat y_\textrm{pseudo}$, along with $y_\textrm{Alt-Inc}$ which acts as a regularizer. %However, we found that the performance of the retrained model is very similar to the performance without the training step, proving that our thresholding operator is indeed successful in not generating false positives (Section \ref{sec: experiments}). 
The unsupervised loss function in Equation~\ref{eq: loss_unsup} can be used to retrain the network on $\hat y_\textrm{pseudo}$, along with $y_\textrm{Alt-Inc}$ which acts as a regularizer. The first row in each experiment shows the results obtained by proposed strategy of pseudo-labeling while the second row shows the results obtained by re-training the generated pseudo labels. However, we found that the performance of the retrained model is very similar to original model, mitigating the need for costly retraining and therefore contributing to the simplicity of the proposed pseudo labeling strategy. %It can be clearly observed that the training step increases the accuracy of the shared classes, thus attenuating the effect of false positives. 

\paragraph{Qualitative Results and Realtime Performance:} We present the qualitative results in Figure \ref{fig:visualisations_altinc}. Our method works well in environments that have dirt roads, absence of clear lane markings, multiple road objects and unstructured traffic. Figures~\ref{fig: boundless_object} and~\ref{fig: boundless_object_result} show that BoMuDANet can recognize auto-rickshaws (boundless category object) reasonably well.

We have included a video demonstration of BoMuDANet in realtime in $6$ diverse traffic videos containing both unconstrained (IDD) as well as structured (CityScapes) environments, along with comparisons. BoMuDANet operates at $2$ fps on IDD and $21$ fps on CityScapes with a model size of $26.5$ million parameters. We refer the reader to the supplementary video.

\begin{table}[t]
% \footnotesize
\centering
% \begin{center}
\resizebox{\columnwidth}{!}{
\begin{tabular}{ l c c c c}
\toprule

% \toprule
\multicolumn{5}{c}{\Tstrut On IDD \Bstrut}\\
\toprule
 Method & Model & \# Size(M$\downarrow$) & mIoU(S$\uparrow$) & mIoU(P$\uparrow$)  \\
 \midrule
 
% \multicolumn{6}{c}{ \Bstrut On IDD}\\
% \hline
\cite{iiith} \Tstrut & ResNet-18 &\textbf{11.70}  & 27.45 & NA \\

\cite{bucher2019zero} & ResNet 101 & 44.50 & 29.20 & 7.90 \\
\cite{bucher2020buda} (UDA) & ResNet 101 & 44.50 & 32.40 & 8.10 \\
\cite{bucher2020buda} (Apt.)& ResNet 101 & 44.50 & 32.70 & 8.60 \\
\cite{bucher2020buda} (Comb.) & ResNet 101 & 44.50 & 37.30 & \textbf{18.50} \\
\textbf{BoMuDANet} & \textbf{DRN-D-38} & \textcolor{blue}{\textbf{26.50}} & \textbf{39.23} & \textcolor{blue}{\textbf{11.85}}\\

\toprule
\toprule 
\multicolumn{5}{c}{\Tstrut On CS \Bstrut}\\
\toprule
 Method & Model & Size(M$\downarrow$) & mIoU(S$\uparrow$) & \# Sources  \\ 
 \midrule

 \cite{chenli}  & ResNet-101 & 44.50 & 39.40 & Single\\
\cite{tsai2018learning} & ResNet-101 & 44.50 & 42.40  & Single\\
 \cite{vu2019advent} & ResNet-101 & 44.50 & 43.10  & Single\\
   \cite{vu2019advent} & ResNet-101 & 44.50 & 43.80  & Single\\
\cite{pan2020unsupervised} & ResNet-101 & 44.50 & 46.30  & Single\\
 \cite{licontent2020} & ResNet-101 & 44.50 & 49.90  & Single\\
 
% \cline{}
% \multicolumn{5}{c}{ \Bstrut Multi-source}\\
% \hline
 \cite{zhao2018adversarial} & VGG-16 & 138.00 & 29.40  & Multi\\
 \cite{zhao2019multisemantic} & VGG-16 & 138.00 & 41.40   & Multi\\
BoMuDANet  &\textbf{DRN-D-38}& \textbf{26.50} & 44.63   & Multi\\
BoMuDANet & ResNet-101 \cite{vu2019advent} & 44.50 & 49.59 & Multi \\
\textbf{BoMuDANet} & ResNet-101 \cite{licontent2020} & 44.50 & \textbf{55.90} & Multi \\

\bottomrule

\end{tabular}
}
\caption{\textbf{Comparison with SOTA:} We compare with the SOTA in both unstructured (IDD) as well as structured (CS) traffic. Higher ($\uparrow$) mIoU and mAcc indicates direction of better performance. \textbf{Bold} indicates best while \textcolor{blue}{\textbf{blue}} indicates second-best. mIoU(S) and mIoU(P) denote the performance on shared/known and private/unknown classes, respectively. \textbf{Conclusion:} Our model is SOTA on IDD by $5.17\%-42.9\%$ and on CS in the multi-source setting by $12.70\%-90.13\%$, with a reduction in model size by upto to $5.2\times$.}

\label{tab:sota}
\vspace{-10pt}
% \end{center}
\end{table}

%VGG 138 M parameters, Deeplab 44.5, DRND 38 26.5

\subsection{Comparsons with SOTA}
\label{subsec: comparison}

\paragraph{In Unstructured Environments (Table~\ref{tab:sota}, On IDD):} In Table \ref{tab:sota} (On IDD), we compare our approach against other unsupervised segmentation methods. ZS3Net~\cite{bucher2019zero} does zero shot semantic segmentation, while~\cite{bucher2020buda} (UDA) and~\cite{bucher2020buda} (Apt.) builds upon ZS3Net for domain adaptation. \cite{bucher2020buda} (Comb.) refers to the combined approach for boundless unsupervised domain adaptation (``BUDA''). It can be clearly observed that our method surpasses all past unsupervised segmentation methods by $5.17\%$ - $34.34\%$ on shared classes, with a much smaller architecture (Table \ref{tab:sota}, model size) which is beneficial for practical autonomous driving real-time applications. 

Our hypothesis for the superiority of BoMuDANet over BUDA is that the latter comprises performance on closed-set classes in order to achieve improved performance on open-set classes \cite{bucher2019zero,bucher2020buda}. In contrast, our method classifies open-set categories \textit{without} sacrificing accuracy on closed-set categories (Table~\ref{tab:open_set}, Figure~\ref{fig:visualisations_altinc}). The decreased performance of BUDA on ``shared'' classes could be due to decreased generalization capabilities of the model when trained on the new classes. We also outperform the semi-supervised method \cite{kalluri2019universal} by $42.9\%$, that uses ground-truth in $100$ samples for supervision. \cite{kalluri2019universal} fails to acknowledge differences between various domains, which leads to a degradation in performance. 

\paragraph{In Structured Environments (Table~\ref{tab:sota}, On CS):} We additionally benchmark BoMuDANet in structured environments, using CS as the target domain and BDD, IDD and GTA as the source domains. Our method is SOTA in the multi-source setting by at least $12.70\%$\textminus$90.13\%$ with a reduction in model size by upto $5.2\times$. Methods with ResNet-101 backbone have an inference time of $156.44$ ms, and models with DRN-D-38 backbone have an inference time of 51.58 ms. On CS, BoMuDA outperforms the corresponding single-source DA baselines by $2.5\% - 21.2\%$ respectively. Furthermore, stronger backbones will help our model benefit accordingly (Table \ref{tab:idd_mainresults} I and III; and second half of Table \ref{tab:sota}). Further, comparison of our network (with corresponding backbones) against single-source baselines reveals that our model is the SOTA (Table \ref{tab:idd_mainresults}, second half of Table \ref{tab:sota}). 

The core step in the approach of~\cite{zhao2019multisemantic} is the use of the CycleGAN \cite{zhu2017unpaired}, which uses images and ground truth from all source domains at every training step. Our multi-source approach, in contrast, is more computationally efficient and requires data only from the ``best source'. Pre-trained single-source adaptation weights can be directly used for the other datasets, thus offsetting the need for images and GT from all source domains. The improvement in our approach comes from individually distilling relevant information from multiple domains as opposed to considering images from all source domains in every iteration.

\subsection{Ablation Studies and Additional Experiments} 

We show the benefits of using multiple sources compared to a single source in Table~\ref{tab:idd_mainresults}. 
% The single source baselines are created by training each source independent from other sources. 
The multi-source model outperforms the corresponding single source baselines by $3.3\% - 54.15 \%$, demonstrating the efficiency of using multiple sources. In boundless DA, we replace the thresholding operator with the KL divergence loss to measure the similarity between the open-set classes and physically similar categories in Table~\ref{tab:open_set}. We demonstrate the iteration wise performance of the Alt-Inc algorithm and a study of tuning the hyperparameters $\lambda_{distil}$ and $\lambda_{unsup}$ in the supplementary material. Additionally, selecting the best source at the pixel-level degrades accuracy by $18.05$ \% due to loss of contextual information. Finally, thresholding on pseudo labels \cite{bucher2020buda} to reduce the number of false positives reduces the mIoU by $4.54$ \%.

\section{User Study}

We conducted a web-based user study to evaluate the visual scene understanding results generated by our approach in unstructured driving environments and evaluate their perceptual benefits compared to prior methods. We also test our hypothesis of classifying open-set objects. These open-set objects correspond to new objects that  our model has never seen before in the training dataset. Finally, we perform an additional study where we highlight the complexity of our task by asking users to rate the level of difficulty of driving in the kinds of scenes evaluated by our ethod.
% test the two major aspects of our method: identification of open-set objects, superior performance using multiple source compared to single-source baselines.

\begin{figure}[t]
    \centering
    % \captionsetup[subfigure]{labelformat=empty}
    \begin{subfigure}[b]{\columnwidth}
    \includegraphics[width = \textwidth]{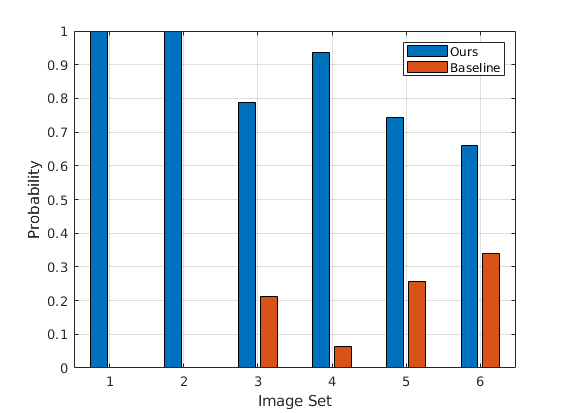}
    \caption{Study $1$: We show the probability of users selecting the result generated by our multi-source approach (blue) on six randomly chosen images of unstructured driving environments from the IDD dataset compared to single-source baselines (red) \cite{tsai2018learning}}
    \end{subfigure}
    \begin{subfigure}[t]{\columnwidth}
    \includegraphics[width = \textwidth]{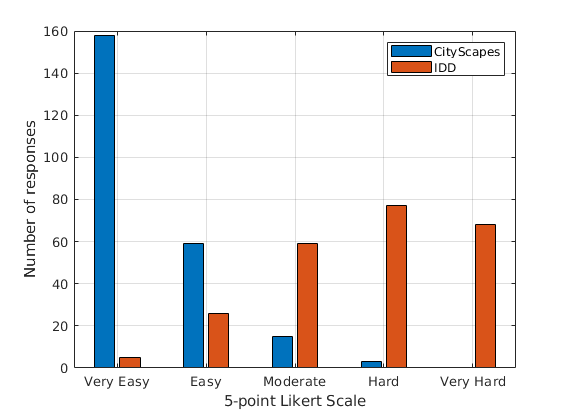}
    \caption{Study 3: We report the number of users corresponding to the $5$-point Likert scale summed over all five sets of images.}
    \end{subfigure}
    \caption{\textbf{User Study Results:} In the first user study, we observe that users are more likely to report our result as the better segmentation compared to state-of-the-art. In the second study, we observe that majority of the users confirm that the IDD traffic scenes are complex and that it would be hard to drive in such environments whereas almost all participants find the driving environments in the CityScapes dataset to be easy.}
    \label{fig:user_study_results}
    % \vspace{-10pt}
\end{figure}

\subsection{Procedure}

The study consists of three sections and is approximately five minutes in duration. In the first section, we show users six comparisons (selected from the dataset randomly) between the results obtained by our unsupervised adaptive multi-source approach versus the state-of-the-art single-source baselines~\cite{tsai2018learning}, along with the corresponding ground truth image. We do not reveal the identity of the methods and swap the order of the results to reduce the bias. We ask users to visually inspect both the results and report the result that best resembles the ground truth image. The aim of this study is to complement and support the quantitative results in Section~\ref{sec: experiments} with human visual feedback.

In the second section, we show users results of our approach on two images containing an open-set and a closed-set object (selected from the dataset randomly) , respectively. We ask the participants to select which result in their view is better. We purposefully avoid defining what constitutes a better segmentation result to  eliminate any bias. Higher probabilities of selecting the open-set object validates our hypothesis of classifying open-set objects.

In the third and final section, we show participants five pairs of images of structured and unstructured driving environments (selected from the dataset randomly) . Following the previous two sections, we do not label the images. Users were asked to rate the difficulty of driving for each image on a $5$-point Likert scale ($1-$ ``Very Easy'' to $5-$ ``Very Hard'').

\subsection{Participants}
$47$ anonymous users participated in our study, recruited over the web. To remove bias related to users with driving experience in a particular traffic environment, we invited participants belonging to regions with traffic that is similar to the traffic found in both CityScapes and the IDD datasets. We do not consider age or gender of the users in our study.

% \divya{Uttaran paper: Fifty participants participated in our study, recruited via web advertisements. To study the demographic diversity, we asked the participants to report their gender and age group. Based on the statistics, we had 16 male and 11 female participants in the age group of 18-24, 15 male and seven female participants in the age group of 25-34, and one participant older than 35 who preferred not to disclose their gender. However, we did not observe any particular pattern of responses based on the demographics.}

\subsection{Evaluation}

\subsubsection{BoMuDANet vs. state-of-the-art}
Each participant picked the segmentation map (out of baseline, and BoMuDANet) that they felt best resembled the ground-truth. We summarize the percentage of responses for each of the $6$ images in Figure \ref{fig:user_study_results}(a), arranged in the order of increasing complexity. All participants agree that BoMuDA performs better than the baseline for image 1 and 2. While images $3-6$ are hard to segment accurately, at least $60-90\%$ of the participants agree that BoMuDA results in a better solution. This, coupled with the fact that participants did not apriori which segmentation maps were generated by the baseline single-source models and BoMuDANet indicates that our model is superior. One possible explanation of participants rating BoMuDANet segmentation maps higher is that BoMuDANet outperforms the single-source baseline by selectively extracting the best features from each source and is thus capable of segmenting dirt roads, persons, motorcycles in a better fashion.

\subsubsection{Open-set analysis}
Define $p$ as the probability that a participant will choose the open-set result (auto-rickshaw) when shown a pair of results generated by our model on an one open-set object and a regular object in the training dataset. We wish to test the hypothesis that either a participant cannot tell the difference between the two results ($p=1/2$) or the open-set result is good enough such that participants choose the same ($p > 1/2$). Thus, we test the following right-tailed hypothesis test:
\begin{equation*}
    H_0: p < \frac{1}{2} \quad \text{against} \quad H_1: p \geq \frac{1}{2}.
\end{equation*}
Let $N_1$ be the number of participants who selected the open-set result as the better segmentation result. Then, $\hat{p} = N_1/N$. We test the above hypothesis with the test statistic $(\hat{p}-p)/\sqrt{(\hat{p}(1-\hat{p}))/N}$, which is asymptotically normal with mean $0$ and standard deviation $1$, under the null hypothesis. We observe in our study that $N_1 = 30$ corresponding to $\hat p = 0.63$ with a p-value of $0.032884$ ($\alpha = 0.05$ significance level). Our result is statistically significant indicating we can reject the null hypothesis ($H_0$) and accept the alternative ($H_1$).

\subsubsection{Complexity of the task}
Each participant was shown 5 pairs of images depicting scenes from CityScapes and IDD (unknown apriori) and asked to rate the difficulty of driving. Therefore, we have a total of 235 responses each, rating the difficulty of driving on a $5$-point scale. We summarize the responses in Figure \ref{fig:user_study_results}(b). About 93\% of the participants reported that it they felt it was easy to drive in CityScapes-like environments, while 63\% of the participants reported that driving in an environment similar to IDD is hard. In contrast, only 14\% of the participants reported that driving in an IDD-like environment is easy and a meagre number of participants felt that navigating through the traffic depicted in CityScapes is hard. This is in line with prior studies \cite{tesla_dirt}, which state that driving in unstructured environments with heterogeneous traffic is difficult.

\section{Conclusion, limitations and future work}
We present a novel learning methods for visual scene understanding in unstructured traffic environment. Our approach consists of a semantic segmentation technique that solves three key aspects of domain adaptation: unsupervised, multi-source and boundless, in unconstrained environments.  We present a novel training routine called Alt-Inc that builds on the ideas of self-training and pseudo-labeling. Alt-Inc is used to selectively distil information from various sources by iterative self-training, in addition to exploiting a chosen best source via domain adaptation. In addition, BoMuDANet can identify unknown objects encountered during the testing phase via a simple pseudo labeling strategy. We highlight the benefits  of our approach in terms of performing accurate segmentation and visual scene understanding in challenging datasets such as IDD. We highlight improved accuracy over prior methods and perform qualitative evaluation based on a user study.

Our approach has some limitations. Our current approach can only recognize new objects by taking advantage of the structural similarities between various objects in road environments. Currently, our model is unable to detect classes like animals and other classes that do not share any similarities with the `known' classes. In addition, the existence of multiple `unknown' objects that share similarities with the same set of `known' classes can cause inter-class confusion. As part of future work, it would be useful to use the pseudo labeling strategy as a prior, and develop a training method that exploits zero-shot learning strategies. The core step in the Alt-Inc algorithm is the selection of a `best source', which can vary from image to image. Our current formulation does not account for this factor during training. In addition, we do not account for variations within the target dataset, where images can have varying levels of similarities with the source datasets. Future work in this area can focus on an importance weighting scheme for a multi-source domain adaptation network that is more robust. We would also like to evaluate our approach in other unstructured and challenging scenarios and integrate with planning and navigation methods.

\bibliographystyle{acm}

\bibliography{references}

\end{document}